\documentclass{article}

\usepackage[dblblindworkshop, final]{neurips_2026}
\workshoptitle{Geometric Distributional Deep Learning}

\usepackage[utf8]{inputenc} 
\usepackage[table]{xcolor}
\usepackage{colortbl}
\usepackage{multirow}
\usepackage{tabularx}
\usepackage{array}
\usepackage{arydshln}
\usepackage{hyperref}
\usepackage{url}
\usepackage{graphicx}
\usepackage{wrapfig}        
\usepackage{algorithm}
\usepackage{algpseudocode}
\usepackage{amsmath, amssymb, amsthm}
\usepackage{subcaption}
\usepackage[T1]{fontenc}    
\usepackage{hyperref}       
\usepackage{url}            
\usepackage{booktabs}       
\usepackage{amsfonts}       
\usepackage{nicefrac}       
\usepackage{microtype}      
\usepackage{xcolor}         
\usepackage{ulem}
\usepackage[most]{tcolorbox}
\tcbuselibrary{skins}

\newtheorem{definition}{Definition}
\newtheorem{lemma}{Lemma}
\newtheorem{proposition}{Proposition}

\definecolor{lightgreen}{RGB}{220, 245, 220}
\definecolor{darkgreen}{rgb}{0.0, 0.5, 0.0}
\definecolor{brg}{HTML}{932A2E}
\hypersetup{colorlinks=true, citecolor=black, linkcolor=brg}
\title{Learning Transferable Representations Across Distributional Shifts via Koopman Operator and Optimal Transport}

\author{%
  \large Abdel Djalil Sad Saoud \\ 
  Université Paris Saclay\\
   CEA-LIST,  F-91120, Palaiseau, France\\
  \texttt{Abdel-Djalil.SADSAOUD@cea.fr} 
  \And
   \large Fred Maurice Ngolè Mboula \\
  Université Paris Saclay\\
   CEA-LIST,  F-91120, Palaiseau, France\\
  \texttt{fred.ngole-mboula@cea.fr} 
    \And
   \large Cédric Gouy-Pailler \\
  Université Paris Saclay\\
   CEA-LIST,  F-91120, Palaiseau, France\\
  \texttt{cedric.gouy-pailler@cea.fr}  
}

\begin{document}

\maketitle

\begin{abstract}
In this paper, we propose a  domain adaptation method that learns cross domain representations by connecting optimal transport and transfer operators theory. We model the source and target distributions as Gaussian mixtures and solve an entropic optimal transport problem between their components, providing a discrete backbone that captures cross-domain correspondences. We then leverage the component densities to lift this backbone into a continuous artificial dynamics designed to preserve semantic class structure, and characterize this dynamics through its Koopman operator.  We motivate how, under favorable connectivity conditions, the slow eigenmodes provide a representation that is stable across domains and naturally defined throughout the ambient space, thereby enabling out-of-sample mapping. Moreover, our operator construction confines the dynamics to an explicit finite-dimensional function space, so that the representation is structurally prescribed by the probabilistic model rather than learned within a generic hypothesis class. Our approach achieves a better performance compared to shallow OT, deep domain adaptation and neural OT baselines on benchmarks spanning audio, signal processing and visual adaptation modalities.
\end{abstract}

\section{Introduction}
Statistical learning theory relies in the assumption that training and test samples are independent and identically distributed. In practice, however, once a model is trained and deployed, the conditions under which inference-time data is acquired often diverge from those of the training set, inducing distributional shifts that alter the statistical structure of the data. For instance, in computer vision, the same visual content may have different appearances due to changes in illumination, camera hardware, or capture conditions. In speech and signal processing, the same underlying phenomenon is represented differently due to differences in sensor characteristics, recording environments, or background noise. Such shifts may cause a model that does well on the \textit{source domain} to degrade unpredictably on the \textit{target domain}, not because the model failed or even overfit the training data but because the statistical structure it relied on no longer holds. This problem is at the center of domain adaptation \citep{pan2010survey}, the problem of finding a learning paradigm that is robust to distributional shift by leveraging knowledge transferred from the domain where it was initially acquired \citep{redko2020survey}. A common assumption in this setting is covariate shift \citep{SHIMODAIRA2000227, courty2017OTDA} where the marginal distributions of inputs are different across domains \textit{i.e.,} $P_s(x) \neq P_t(x)$, but the conditional distribution of labels given inputs is the same, $P_s(y\mid x) = P_t(y\mid x)$ \textit{i.e.,} the task is preserved. To this end, Optimal Transport (OT) \citep{villani2009optimal, courty2017OTDA} provides a theoretically grounded way to compare and align probability distributions. OT aims to find the minimum effort strategy to transport samples from one distribution to another and computes a transport plan that can be used to map the source samples toward the target distribution, where \cite{courty2017OTDA} proposed an approximation of the Monge mapping by using the so-called barycentric mapping to do so. However, the OT plan is sensitive to regularization \citep{cuturi2013sinkhorn}, and mapping source samples to the target domain this way may produce biased domain alignment \citep{sadsaoud2026beyond}. Moreover, most shallow OT methods do not provide \textit{out-of-sample mapping} \citep{seguy2018large}, in the sense that the transport plan is computed exclusively between the discrete measures \citep{montesuma2025optimal}. Hence, any newly observed point not considered in the original OT problem cannot be mapped without recomputing the OT plan from scratch, which leaves no guaranty that the classifier will correctly predict such points at inference.


\textbf{Contributions.} Our contributions are threefold. 
First, we reinterpret domain shift as an artificial dynamics between the source and target domains, whose transitions are designed to preserve the semantic class structure. We construct its continuous transition law from Gaussian mixture models and entropic OT, and characterize the resulting dynamics through its Koopman operator. Second, we provide conditions under which its slow eigenmodes are stable across domains within classes, enabling classifier training from labeled source data alone. Being defined throughout the input space, they also naturally provide out-of-sample mapping for unseen source or target points. Third, we show that, despite the continuous state space, the spectral dynamics of the Koopman operator lies in a finite-dimensional function space explicitly determined by the mixture model, reducing eigenmode computation to a finite-dimensional fixed-point problem. We jointly learn the slow eigenmodes end-to-end with a shallow classifier, outperforming shallow OT, deep domain adaptation, and neural OT baselines across benchmarks spanning multiple modalities.


\textbf{Notation.} Scalars, vectors, and matrices are denoted by lowercase $a$, boldface lowercase $\mathbf{a}$, and boldface uppercase $\mathbf{A}$ respectively. For a matrix $\mathbf{A}$, $\mathbf{A^\top}$ denotes its transpose and $\mathbf{A^{-1}}$ its inverse when $\mathbf{A}$ is invertible. The canonical inner product between two vectors is $\mathbf{\langle u, v\rangle} = \sum_i u_i v_i$, with the Frobenius inner product $\langle \mathbf{U, V}\rangle_F = \mathrm{Tr}(\mathbf{U^\top V}) $ for matrices. We denote by $\|\cdot\|_2$ the Euclidean norm and $\|\cdot\|_F$ the Frobenius norm. The identity matrix of size $n$ is $\mathbf{I_n}$. We write $\mathbf{1}_n = (1, 1,\dots,1)^\top \in \mathbb{R}^n$ for the vector of all ones, so that for $\mathbf{A}\in\mathbb{R}^{n\times m}$, $\mathbf{A}\mathbf{1}_m\in\mathbb{R}^n$ and $\mathbf{A^\top}\mathbf{1}_n\in\mathbb{R}^m$ denote its row and column sums respectively. $\delta_x$ denotes the Dirac measure at point $x$, \textit{s.t.} a discrete measure $\mu$ can be written as $\mu = \sum_{i=1}^n a_i\delta_{x_i}$ with $a_i \geq 0$ and $\sum_{i=1}^n a_i = 1$.  $\mathcal{M}^1_+(\mathcal{X})$ denotes the set of probability distributions over a metric space $\mathcal{X}$. We use $\mathrm{Pr}(\cdot)$ to denote probabilities and $\mathbb{E}[\cdot]$ for expectations.
\section{Related Work}\label{sec:related_work}

\textbf{Deep Domain Adaptation. }Deep domain adaptation methods learn domain-invariant representations either by minimizing a statistical discrepancy between source and target features in a shared latent space via MMD-based adaptation layer and domain confusion loss \cite{tzeng2014deep}, multi-kernel MMD across all task-specific layers embedded in a reproducing kernel Hilbert space \cite{long2015learning}, and second-order statistic alignment \cite{sun2016return}, or via adversarial training methods, where a gradient reversal layer \cite{ganin2016domain}, conditional domain discriminator and classifier predictions \cite{long2018conditional}, or two-classifier discrepancy maximization and a near support feature generator \cite{saito2018maximum} are used to produce domain-invariant features. 

\textbf{Optimal Transport. } Optimal transport has been used both to design domain adaptation algorithms and to theoretically analyze the problem \cite{courty2017OTDA, redko2017theoretical}. The foundational approach of \cite{courty2017OTDA} uses the Kantorovich relaxation to compute a regularized transport plan between source and target empirical distributions, and defines the  barycentric mapping of source samples toward the target domain. In \cite{perrot2016mapping}, the coupling and an approximation of the transport map are jointly learned through a convex formulation, enabling out-of-sample generalization of the mapping. Linear Monge mapping estimation, together with concentration limits and generalization guarantees for OT-based domain adaptation, were studied in \cite{flamary2019concentration}. JDOT \cite{courty2017joint} and DeepJDOT \cite{damodaran2018deepjdot} extended the framework by simultaneously optimizing the transport coupling and a prediction function $f$ to align the full joint distribution $P(X,f(X))$. Beyond empirical distributions, a Wasserstein-type distance on the space of Gaussian mixture models was introduced in \cite{delon2020wasserstein} by restricting the OT coupling to Gaussian mixtures, showing that this continuous formulation reduces to an efficient discrete problem between mixture components. Building on this, two unsupervised domain adaptation strategies were proposed in \cite{montesuma2025optimal} based on label propagation via the GMMOT plan, and sample mapping between domains. However, as acknowledged by the authors, this approach rely on a sufficiently well-separated multi-modal structure, which may be difficult to satisfy in high dimensions.  Beyond discrete solvers, a related OT literature addresses the limitation of transport plan being defined only between finite empirical measures by learning continuous transport maps. As in \cite{seguy2018large}, a two-step approach was proposed, by first learning a stochastic dual OT plan then estimating a Monge map as a neural network approximating its barycentric projection, providing out-of-sample extension. In \cite{makkuva2020optimal}, the optimal transport map is parametrized  as the gradient of a convex function, and the optimal Kantorovich potential is learned via minimax optimization over input-convex neural networks (ICNNs), which are also used in \cite{taghvaei20192, 
korotin2021wasserstein}. A unified framework for learning continuous OT maps and plans was proposed in \cite{korotin2023neural}, for both strong and weak transport costs, with universal approximation guarantees. The latter methods provide out-of-sample mapping, but are primarily designed for generative modeling and unpaired image-to-image translation rather than domain adaptation.

\textbf{Spectral Domain Adaptation. }A related line of work exploits spectral structures for domain adaptation, either via Laplacian-based spectral regularizer \cite{xiao2023spa} or label propagation \cite{el2021label}. Closer to our work, SeOT \cite{sadsaoud2026beyond} constructs a weighted adjacency matrix from OT plans between discrete measures and extracts domain-invariant representations via spectral embedding with Arnoldi solver, but remains limited to a discrete formulation with no out-of-sample extension. To the best of our knowledge, no existing work combines Koopman operator theory with optimal transport to derive a continuous domain adaptation formulation that supports out-of-sample extension.

\section{Background}\label{sec:background}

\subsection{Optimal Transport}\label{sec:background_ot} Optimal transport (OT) originates from the problem first posed by Monge \cite{monge1781memoire} of displacing mass at minimum total effort. Its modern measure-theoretic formulation, due to Kantorovich \cite{kantorovich1942translocation}, formulates this as an optimization problem over joint probability distributions, and has since then become a powerful tool in machine learning \citep{villani2009optimal,peyre2019computational, montesuma2024recent}. 

\begin{definition}[Kantorovich Relaxation]
Let $\mu$ and $\nu$ be two probability distributions over a metric space $\mathcal{X}$, 
and let $c : \mathcal{X} \times \mathcal{X} \to \mathbb{R}_+$ be a ground cost function measuring the effort of transporting a unit of mass from $x$ to $x'$. The Kantorovich optimal transport problem seeks the coupling of minimum transport cost:
\begin{equation}\label{eq:kantorovich}
    \pi^* = \underset{\pi \in \Pi(\mu,\nu)}{\mathrm{arg\,inf}} 
    \int_{\mathcal{X}\times\mathcal{X}} c(x,x')\,d\pi(x,x'),
\end{equation}
where $\Pi(\mu, \nu) = \left\{\pi \in \mathcal{M}^1_+(\mathcal{X}\times\mathcal{X}) 
\;\middle|\; \int_{\mathcal{X}}\pi(\cdot, x')\,dx' = \mu, \;\int_{\mathcal{X}}\pi(x, \cdot)\,dx = \nu \right\}$ is the set of joint distributions over $\mathcal{X}\times\mathcal{X}$ whose marginals 
are $\mu$ and $\nu$ respectively.
\end{definition}

In the domain adaptation setting, $\mu$ and $\nu$ are referred to as the source and target measures $\mu_s$ and $\mu_t$ respectively, both supported on a shared feature space $\mathcal{X}_s = \mathcal{X}_t = \mathcal{X} \subseteq \mathbb{R}^d$ (homogeneous domain adaptation), and we adopt this notation throughout. The formulation in \eqref{eq:kantorovich} is an infinite-dimensional problem \citep{peyre2019computational, montesuma2025optimal}. In practice, one only has access to finite sets of source and target observations $\boldsymbol{X_s} = \{\mathbf{x_s}^i\}_{i=1}^{n_s}$ and $\boldsymbol{X_t} = \{\mathbf{x_t}^j\}_{j=1}^{n_t}$, respectively. A natural way to represent these observations as probability distributions is through the empirical measures $\mu_s = \sum_{i=1}^{n_s}a_i\,\delta_{\mathbf{x_s}^i}$ and $\mu_t = \sum_{j=1}^{n_t}b_j\,\delta_{\mathbf{x_t}^j}$, with weights $a_i,b_j\ge 0$ satisfying $\sum_{i=1}^{n_s}a_i=\sum_{j=1}^{n_t}b_j=1$, supported on $\boldsymbol{X_s}$ and $\boldsymbol{X_t}$ respectively. Under this discretization, \eqref{eq:kantorovich} reduces to the following finite linear program:
\begin{equation}\label{eq:discrete_ot}
    \boldsymbol{\pi}^* = \underset{\boldsymbol{\pi} \in \Pi(\mu_s, \mu_t)}{\mathrm{arg\,min}}\; 
    \langle \boldsymbol{\pi, C} \rangle_F = \underset{\pi \in \Pi(\mu_s, \mu_t)}{\mathrm{arg\,min}} 
    \sum_{i=1}^{n_s}\sum_{j=1}^{n_t} \pi_{ij}\, C_{ij},
\end{equation}
where $\mathbf{C}\in\mathbb{R}^{n_s\times n_t}$ is the cost matrix with entries $C_{ij} = c(\mathbf{x_s}^i, \mathbf{x_t}^j) = \|\mathbf{x_s}^i - \mathbf{x_t}^j\|_2^2$ \textit{i.e., here, we consider the squared Euclidean distance}, and
 $\Pi(\mu_s, \mu_t) = \{\boldsymbol{\pi} \in \mathbb{R}_+^{n_s \times n_t} 
\mid \textstyle\sum_{j=1}^{n_t}\pi_{ij} = a_i\; \forall i ,\ \sum_{i=1}^{n_s}\pi_{ij} = b_j \; \forall j\}.
$ This problem is convex in $\boldsymbol{\pi}$ and admits at least one solution, yet solving it exactly via linear programming \citep{dantzig2016linear} scales in $\mathcal{O}(n^3\log n)$ when $n_s = n_t = n$, which is computationally expensive at the scales of machine learning applications \citep{peyre2019computational}. To alleviate this computational burden, adding an entropic regularization term was proposed in \cite{cuturi2013sinkhorn}. In its continuous form, it reads:
\begin{equation}\label{eq:entropic_ot_continuou}
    \pi^*_\varepsilon \;=\; \underset{\pi \in \Pi(\mu_s, \mu_t)}{\mathrm{arg\,inf}}\;
    \int_{\mathcal{X}\times\mathcal{X}} c(x, x')\,d\pi(x,x')
    \;+\; \varepsilon\,
    \underbrace{
    \int_{\mathcal{X}\times\mathcal{X}}
    \log\!\left(\frac{d\pi(x,x')}{d(\mu_s\otimes\mu_t)(x,x')}\right) d\pi(x,x')
    }_{\mathrm{KL}(\pi\,\mid\mid\mu_s\otimes\mu_t)},
\end{equation}
where $\varepsilon > 0$ controls the regularization strength, and KL$(\pi \mid\mid \mu_s\otimes\mu_t)$ is the Kullback-Leibler divergence between the candidate coupling $\pi$ and the product measure $\mu_s\otimes\mu_t$. The discrete entropic optimal transport problem reads:
\begin{equation}\label{eq:entropic_ot_discrete}
    \boldsymbol{\pi}^*_\varepsilon \;=\;
    \underset{\boldsymbol{\pi} \in \Pi(\mu_s, \mu_t)}{\mathrm{arg\,min}}\;
    \langle \boldsymbol{\pi}, \mathbf{C}\rangle_F \;+\; \varepsilon\, \text{\textbf{KL}}(\boldsymbol{\pi} \mid\mid \mu_s\otimes\mu_t),
    \quad
    \text{\textbf{KL}}(\boldsymbol{\pi} \mid\mid \mu_s\otimes\mu_t) \;\overset{\text{def.}}{=}\; \!\sum_{i,j}\pi_{ij}\log \bigl(\frac{\pi_{ij}}{a_i b_j}\bigr).
\end{equation}
The appeal of~\eqref{eq:entropic_ot_discrete} is that the \textbf{KL} term makes the objective strictly convex in $\boldsymbol{\pi}$, ensuring a unique minimizer that can be obtained through the iterative matrix-scaling procedure of \cite{sinkhorn1967concerning}, where each iteration reduces to matrix-vector products, bringing the cost to $\mathcal{O}(n^2)$ \citep{peyre2019computational}. Note that  Entropic OT induces a smoothing in the transport plan as $\varepsilon$ grows. One recovers a maximum-entropy solution of~\eqref{eq:discrete_ot} as $\varepsilon \to 0$, and recovers the independent coupling $\boldsymbol{\pi}^*_\varepsilon \to \mathbf{a} \otimes \mathbf{b} = \mathbf{a} \mathbf{b}^\top$ as $\varepsilon \to +\infty$ \cite{peyre2019computational}. 

\textbf{Gaussian Mixture Optimal Transport.}
When both $\mu_s$ and $\mu_t$ are Gaussian, i.e., $\mu_s = \mathcal{N}(\mathbf{m}^s, \boldsymbol{\Sigma}^s)$ and $\mu_t = \mathcal{N}(\mathbf{m}^t, \boldsymbol{\Sigma}^t)$, the OT problem~\eqref{eq:kantorovich} admits a closed-form solution, the Bures distance \citep{takatsu2011wasserstein}:
\begin{equation}\label{eq:bures_distance}
    \mathcal{W}_2^2(\mu_s, \mu_t) \;=\; \|\mathbf{m}^s - \mathbf{m}^t\|_2^2
    \;+\; \mathrm{tr}\!\left(\boldsymbol{\Sigma}^s + \boldsymbol{\Sigma}^t -
    2\bigl((\boldsymbol{\Sigma}^s)^{1/2}\,\boldsymbol{\Sigma}^t\,
    (\boldsymbol{\Sigma}^s)^{1/2}\bigr)^{1/2}\right).
\end{equation}
Furthermore, provided $\boldsymbol{\Sigma^s}$ is non-singular, the corresponding 
optimal transport map takes the affine form \citep{delon2020wasserstein}:
{\begin{equation}\label{eq:optimal_map_gaussian}
    \forall x \in \mathbb{R}^d,\;\; T(x) \;= \mathbf{m}^t \;+\; (\boldsymbol{\Sigma}^s)^{-1} \bigl(\boldsymbol{\Sigma}^s\,\boldsymbol{\Sigma}^t\bigr)^{1/2} 
    (x - \mathbf{m}^s).
\end{equation}}
\hspace{-0.15cm} This setup was extended in \cite{chen2018optimal} and \cite{delon2020wasserstein} to Gaussian mixtures denoted by $P_S = \sum_{k_1=1}^{K_S} \alpha_{Sk_1}\,P_{Sk_1}$ and $P_T = \sum_{k_2=1}^{K_T} \alpha_{Tk_2}\,P_{Tk_2}$, where $P_{Sk_1} = \mathcal{N}(\mathbf{m}_{Sk_1}, \boldsymbol{\Sigma}_{Sk_1})$, $P_{Sk_1} = \mathcal{N}(\mathbf{m}_{Tk_2}, \boldsymbol{\Sigma}_{Tk_2})$, and  $\boldsymbol{\alpha}_S \in \Delta^{K_S}$, $\boldsymbol{\alpha}_T \in \Delta^{K_T}$ \textit{s.t.} $\Delta^{K} = \{\alpha \in \mathbb{R}_+^K \mid \sum_{k=1}^K 
\alpha_k = 1\}$. Since the optimal coupling between two mixtures is generally not itself a mixture, the feasible set is restricted to couplings in $\mathrm{GMM}_{2d}$, and the problem reduces exactly to discrete OT between mixture components:
\begin{equation}\label{eq:mw2_discrete}
    \mathcal{MW}_2^2 \;=\; \underset{\boldsymbol{\pi} \in \Pi(\boldsymbol{\alpha}_S,
    \boldsymbol{\alpha}_T)}{\mathrm{min}}\;
    \sum_{k_1=1}^{K_S}\sum_{k_2=1}^{K_T}
    \pi_{k_1,k_2}\,\mathcal{W}_2^2(P_{Sk_1}, P_{Tk_2}),
\end{equation}
with $\Pi(\alpha_S, \alpha_T) = \{\boldsymbol{\pi} \in \mathbb{R}_+^{K_S \times K_T} \mid \boldsymbol{\pi} \mathbf{1}_{K_T} = \boldsymbol{\alpha}_S,\; \boldsymbol{\pi}^\top \mathbf{1}_{K_S} = \boldsymbol{\alpha}_T\}$ and each cost entry given by the Bures distance~\eqref{eq:bures_distance}. This collapses an infinite-dimensional problem into a discrete OT of size $K_S \times K_T$, far smaller than the empirical $n_s \times n_t$ problem~\eqref{eq:discrete_ot}, while retaining the full mixture structure and yielding an analytically tractable coupling \citep{delon2020wasserstein, montesuma2025optimal}. The optimal \textit{GMMOT} coupling is then defined as $\gamma_{\tiny \text{GMMOT}}^*(x,y) = \sum_{k_1, k_2}\pi_{k_1 k_2} P_{Sk_1}(x) \delta_{y=T_{k_1 k_2}(x)}$, where $T_{k_1, k_2}$ is the $\mathcal{W}_2$-optimal map between source component $P_{Sk_1}$ and target component $P_{Tk_2}$.


\subsection{Koopman Operator and Spectral Embedding}\label{sec:transfer operators}
Spectral embedding is a technique that represents the vertices of a graph in a low-dimensional Euclidean space where geometric proximity reflects the connectivity structure of the graph. In its discrete formulation, given a finite weighted undirected graph $\mathcal{G} = (V, E, \mathbf{A})$ with adjacency matrix $\mathbf{A} \in \mathbb{R}^{n \times n}$ and degree matrix $\mathbf{D} = \operatorname{diag}(\mathbf{A} \mathbf{1}_n)$, one constructs the random-walk normalized Laplacian $\mathbf{L}_{\mathrm{rw}} = \mathbf{I - D^{-1} A}$ and 
takes the eigenvectors associated with its smallest \textit{nonzero} eigenvalues as coordinates, which encode the cluster structure of the graph. For a 
complete treatment of the discrete formulation, we refer the reader to \citep{aggarwal2014data}. Since our framework relies on a continuous formulation, we adopt an equivalent viewpoint based on \textit{transfer operators}, which extends spectral embedding to random-walk processes on continuous state spaces and studies their long-time behavior \citep{klus2024transfer}.

\begin{definition}[Koopman Operator]
Let $f:\mathbb{R}^d \rightarrow \mathbb{R}$ be an observable and let $p(x,y)$ be the transition kernel of a Markov process on $\mathbb{R}^d$.
The Koopman operator $\mathcal{K}$ is the linear operator $\mathcal{K} : L^2(\mathbb{R}^d) \longrightarrow L^2(\mathbb{R}^d)$ defined by:
\begin{equation}\label{koopman_def}
    (\mathcal{K}f)(x) = \int_{\mathbb{R}^d} p(x,y)\, f(y)\, dy.
\end{equation}
The operator $\mathcal{K}$ describes the evolution of observables under the
underlying stochastic dynamics, equivalently along the random walk on the
associated graph $\mathcal{G}$.
\end{definition}
\begin{tcolorbox}[colback=gray!8, colframe=gray!35, boxrule=0.2pt, arc=2pt, left=1pt, right=1pt, top=1pt, bottom=1pt, before skip=2pt, after skip=4pt]
\begin{lemma}[Spectral properties under reversibility]\label{lem:koopman_spectrum}
Let $p(x,y)$ be the transition kernel of a Markov process on $\mathbb{R}^d$ 
that is reversible with respect to an invariant density $\mu$, i.e., $\mu(x)\,p(x,y) = \mu(y)\,p(y,x)$ for almost every $(x,y) \in \mathbb{R}^d \times \mathbb{R}^d$. Then the Koopman operator $\mathcal{K}: L^2(\mathbb{R}^d, 
\mu) \to L^2(\mathbb{R}^d, \mu)$ defined in~\eqref{koopman_def} is self-adjoint with real-valued spectrum contained in $[-1,1]$
\citep[Lemma~2.17 and Proposition~A.1]{klus2024transfer}. The eigenvalue $\lambda = 1$ is always attained with the constant function 
$\mathbf{1}$ as eigenfunction, and its multiplicity equals the number of connected components of the graph induced by $p$ \citep{levin2017markov}.
\end{lemma}
\end{tcolorbox}
Under \autoref{lem:koopman_spectrum}, the eigenfunctions of $\mathcal{K}$ encode the long-term behavior of the random walk, where the subdominant eigenvalues $\lambda_\ell \lesssim 1$ correspond to the eigenfunctions $f_\ell$ that are 
nearly invariant under the dynamics, \textit{i.e.,} $\mathcal{K} f_\ell \approx f_\ell$. These near-invariant eigenfunctions identify  \underline{\textit{metastable sets}} of the walk, \textit{i.e.,} regions of the state space within which a random walker remains for long periods before transitioning out, which are the continuous analogue of clusters in the discrete case. The number of dominant eigenvalues close to $1$, separated from the rest of the spectrum by a clear spectral gap, indicates the number of such metastable sets, and their associated eigenfunctions serve as the spectral embedding of the underlying process \citep{klus2024transfer}.
\section{Domain Dynamics}\label{sec:method}
In this section, we present our method, \textbf{Domain Dynamics (DoDyn)}, for learning representations of samples that are stable across  distributional shifts via Koopman operator and Gaussian mixture OT.
\subsection{Lifting Empirical Distributions to Gaussian Mixtures}\label{sec:emp2gaussian}
The starting point is the discrete spectral embedding induced by the optimal-transport
bipartite graph in~\citep{sadsaoud2026beyond}, where its transport weights can be normalized into transition
probabilities between source and
target samples. This discrete construction, however, only specifies connectivity between a finite set of source and target samples, which leaves unseen points outside this structure. What we want instead is to spread this discrete connectivity over the ambient space, in a way that keeps it concentrated within each sematic class. For this purpose, instead of the discrete empirical measures, we model source and target samples as Gaussian Mixtures. Let   $P_{Sk_1} = \mathcal{N}(\mathbf{m}_{Sk_1}, \boldsymbol{\Sigma}_{Sk_1})$, $k_1=1,\ldots,K_S$, and  $P_{Tk_2} = \mathcal{N}(\mathbf{m}_{Tk_2}, \boldsymbol{\Sigma}_{Tk_2})$, $k_2=1,\ldots,K_T$, be the source and target mixture components, with means $\mathbf{m}$ and covariances $\mathbf{\Sigma}$. The resulting source and target densities are then the Gaussian mixtures $P_S(x_s) = \sum_{k_1=1}^{K_S}\alpha_{Sk_1}\,P_{Sk_1}(x_s)$ and $P_T(x_t) = \sum_{k_2=1}^{K_T}\alpha_{Tk_2}\,P_{Tk_2}(x_t)$. Using this representation, we solve an  entropy-regularized optimal transport problem between the mixtures components, resulting in a transport plan which serves as the discrete cross-domain connectivity backbone. We define the \textit{continuous coupling} to be:
\begin{equation}\label{continuous_coupling}
\pi_{\varepsilon}^*(x_s, x_t) \;\overset{\text{def.}}{=}\; 
\sum_{k_1=1}^{K_S}\sum_{k_2=1}^{K_T} 
\pi_{k_1,k_2}\,P_{Sk_1}(x_s)\,P_{Tk_2}(x_t).
\end{equation}
\begin{tcolorbox}[colback=gray!8, colframe=gray!35, boxrule=0.2pt, arc=2pt, left=1pt, right=1pt, top=1pt, bottom=1pt, before skip=2pt, after skip=4pt]
\begin{lemma}[Marginals of the continuous coupling]
\label{lem:marginals}
$\pi_{\varepsilon}^*(x_s, x_t)$ is a valid continuous joint distribution whose marginals recover exactly $P_S$ and $P_T$, which guarantees that the lifted GMM representations are preserved through the coupling (proof in Appendix~\ref{App:proof_marginals}).
\end{lemma}
\end{tcolorbox}
One might question the choice of independent-product lifting rather than the $\gamma_{\small\text{GMMOT}}^*(x,y)$ of  \cite{delon2020wasserstein}. The reason is that, for a fixed $x$, $\gamma_{\text{GMMOT}}^*$ is supported on finitely many deterministic points, so from any $x_0$ the reachable
set is at most countable and thus has Lebesgue measure zero. Therefore, it does not provide the volumetric within-class
mixing required by the present metastable dynamics. The independent-product lifting~(\ref{continuous_coupling}) instead spreads
each active component connection over the Cartesian product of the two component densities, so that whenever $\pi_{k_1 k_2}>0$ the connection induces a continuous interaction between the regions carrying mass under $P_{Sk_1}$ and $P_{Tk_2}$ rather than a deterministic map. We defer the detailed explanation and proof to  Appendix~\ref{app:reachability}.

\subsection{Markov Dynamics and Koopman Eigenfunctions}

We adapt the graph formulation in \citep{sadsaoud2026beyond} to the 
continuous setting using $\pi_\varepsilon^*(x,y)$ to define a continuous analogue of a weighted undirected graph. The adjacency and degree functions are given by:
\begin{equation}\label{adjacency_degree}
\resizebox{0.9\linewidth}{!}{$\displaystyle
\mathcal{A}(x, y) = \frac{1}{2}[\pi_{\varepsilon}^*(x, y) + \pi_{\varepsilon}^*(y, x)], \quad d(x) = \int_{\mathbb{R}^d}\mathcal{A}(x, y)\,dy = \frac{1}{2}\Big[ \underbrace{\int_{\mathbb{R}^d}\pi_{\varepsilon}^*(x, y)\,dy}_{P_S(x)} + \underbrace{\int_{\mathbb{R}^d}\pi_{\varepsilon}^*(y, x)\,dy}_{P_T(x)}\Big],
$}
\end{equation}
where the marginal identities follow from \autoref{lem:marginals}. This formulation defines an artificial Markov chain on $\mathbb{R}^d$, with the probability of transitioning from $x$ to $y$ given by the transition kernel $p(x, y) \;=\; {\mathcal{A}(x, y)}/{d(x)}$. The symmetrization in~\eqref{adjacency_degree} removes the need to assign an arbitrary domain label to an intermediate state and makes the component-level cross-domain connectivity bidirectional, which is a continuous analogue of an undirected affinity graph. By \autoref{lem:koopman_spectrum} it follows that $\lambda = 1$ is a simple eigenvalue of the associated Koopman operator attained only by the constant eigenfunction $\mathbf{1}$. However, our interest is in metastable, class-discriminative coordinates that are therefore not carried by this trivial eigenpair, but by the subdominant eigenpairs $(\lambda_\ell, f_\ell)$, \textit{s.t.} $\lambda_\ell \lesssim  1$. Hence, substituting 
$p(x,y)$ into \eqref{koopman_def}, the eigenproblem $\mathcal{K}f_\ell = \lambda _\ell f_\ell$ takes the form:
\begin{subequations}\label{koopman_ot}
\begin{align}
\mathcal{K}f_{\textcolor{black}{\ell}}(x) &= \int_{\mathbb{R}^d} f_{\textcolor{black}{\ell}}(y)\, 
\frac{\mathcal{A}(x, y)}{d(x)}\, dy 
\label{koopman_a}\\
&= \frac{1}{P_S(x) + P_T(x)} \int_{\mathbb{R}^d} f_{\textcolor{black}{\ell}}(y)
\Bigl[\pi_\varepsilon^*(x, y) + \pi_\varepsilon^*(y, x)
\Bigr]\, dy \;=\; \textcolor{black}{\lambda_\ell}\, f_{\textcolor{black}{\ell}}(x).
\label{koopman_b}
\end{align}
\end{subequations}
\begin{tcolorbox}[colback=gray!8, colframe=gray!35, boxrule=0.2pt, arc=2pt, left=1pt, right=1pt, top=1pt, bottom=1pt, before skip=2pt, after skip=4pt]
\begin{proposition}[Expansion using component expectations]\label{propo:expansion finite}
Let $\theta_{Sk_1} = \int_{\mathbb{R}^d} f_{\textcolor{black}{\ell}}(y)\,P_{Sk_1}(y)\,dy$ 
and $\theta_{Tk_2} = \int_{\mathbb{R}^d} f_{\textcolor{black}{\ell}}(y)\,P_{Tk_2}(y)\,dy$ 
be the component expectations of $f_{\textcolor{black}{\ell}}$ under the source and target Gaussian components respectively, and 
$g_{Sk_1}(x) = \frac{P_{Sk_1}(x)}{P_S(x)+P_T(x)}$, 
$g_{Tk_2}(x) = \frac{P_{Tk_2}(x)}{P_S(x)+P_T(x)}$.
Then, for $\lambda_\ell \neq 0$, equation \eqref{koopman_b} admits the following closed-form fixed-point equation:
\begin{equation}
    \hat{f}_{{\ell}}(x) = \lambda_\ell f_{{\ell}}(x) = \left[\sum_{k_1=1}^{K_s} g_{Sk_1}(x)\,(\pi\theta_T)_{k_1}
    + \sum_{k_2=1}^{K_t} g_{Tk_2}(x)\,(\pi^\top\theta_S)_{k_2}\right].
    \label{eq:f_g_form_bis}
\end{equation}
\end{proposition}
\end{tcolorbox}
According to \autoref{propo:expansion finite}, the infinite-dimensional eigenproblem in \eqref{koopman_ot} admits an exact 
finite-dimensional reduction. It is not necessary to find $f$ in the whole $L^2(\mathbb{R}^d)$ space, but it suffices to find the $K_S + K_T$ component expectations $\{\boldsymbol{\theta}_{Sk_1}\}_{k_1=1}^{K_S}$ and $\{\boldsymbol{\theta}_{Tk_2}\}_{k_2=1}^{K_T}$, from which $f$ is reconstructed pointwise through \eqref{eq:f_g_form_bis}. The proof is given in Appendix \ref{App:proof_expansion finite}.

\textbf{A Finite-Dimensional Fixed-Point Formulation.}
While \autoref{propo:expansion finite} expresses $f$ in closed form, the component expectations $\boldsymbol{\theta_S}$ and $\boldsymbol{\theta_T}$ 
are themselves defined as integrals of $f$. We now show that this circularity resolves into 
an explicit finite-dimensional fixed-point system. For $i \in \{1,\dots,K_S\}$ (resp. $j \in\{1,\dots,K_T\}$), by substituting \eqref{eq:f_g_form_bis} into $\theta_{Si} = \int_{\mathbb{R}^d} P_{Si}(x)f(x)\,dx$ (resp. $\theta_{Tj} = \int_{\mathbb{R}^d} P_{Tj}(x)f(x)\,dx$) and collecting terms, we introduce the \textit{overlap matrices} $\mathbf{M_{SS}}$, $\mathbf{M_{ST}}$, $\mathbf{M_{TS}}$, and $\mathbf{M_{TT}}$, corresponding respectively to the source-source, source-target, target-source, and target-target interactions, defined as:
\begin{subequations}
\begin{equation}
    M_{SS}[i,k_1] = \int_{\mathbb{R}^d} \frac{P_{Si}(x)P_{Sk_1}(x)}{P_S(x)+P_T(x)}\,dx,
    \quad
    M_{ST}[i,k_2] = \int_{\mathbb{R}^d} \frac{P_{Si}(x)P_{Tk_2}(x)}{P_S(x)+P_T(x)}\,dx,
\label{eq:MSS_MST}
\end{equation}
\begin{equation}
    M_{TS}[j,k_1] = \int_{\mathbb{R}^d} \frac{P_{Tj}(x)P_{Sk_1}(x)}{P_S(x)+P_T(x)}\,dx,
    \quad
    M_{TT}[j,k_2] = \int_{\mathbb{R}^d} \frac{P_{Tj}(x)P_{Tk_2}(x)}{P_S(x)+P_T(x)}\,dx.
\label{eq:MTS_MTT}
\end{equation}
\end{subequations}
In terms of these matrices, the fixed-point equations for $\theta_{Si}$ and $\theta_{Tj}$ written in vector form gives the following \textit{finite-dimensional fixed point system}:
\begin{equation}
\begin{aligned}
\lambda_\ell\,\boldsymbol{\theta_S} &= \mathbf{M}_{SS}\,\boldsymbol{\pi}\,\boldsymbol{\theta}_T +\mathbf{M}_{ST}\,\boldsymbol{\pi}^\top\boldsymbol{\theta}_S,\\
\lambda_\ell\,\boldsymbol{\theta_T} &= \mathbf{M}_{TS}\,\boldsymbol{\pi}\,\boldsymbol{\theta}_T 
+ \mathbf{M}_{TT}\,\boldsymbol{\pi}^\top\boldsymbol{\theta}_S,
\end{aligned}
\quad \boldsymbol{\Theta} = 
[\boldsymbol{\theta}_S^\top, \boldsymbol{\theta}_T^\top]^\top, \quad
\lambda_\ell\,\boldsymbol{\Theta} = 
\begin{bmatrix}
\mathbf{M}_{ST}\boldsymbol{\pi}^\top & \mathbf{M}_{SS}\boldsymbol{\pi}\\
\mathbf{M}_{TT}\boldsymbol{\pi}^\top & \mathbf{M}_{TS}\boldsymbol{\pi}
\end{bmatrix}\boldsymbol{\Theta}.
\label{eq:theta_fP_System_main}
\end{equation}
Thus $\Theta$ is an eigenvector of the block operator in \eqref{eq:theta_fP_System_main} associated with a subdominant eigenvalue $\lambda_\ell \lesssim 1$. Once $\boldsymbol{\theta_S}$ and $\boldsymbol{\theta_T}$ are known, the Koopman eigenfunction is reconstructed pointwise according to \ref{eq:f_g_form_bis}. See Appendix \ref{App:theta_derivation} for the complete derivation of~\eqref{eq:theta_fP_System_main}.
\begin{tcolorbox}[colback=gray!8, colframe=gray!35, boxrule=0.2pt, arc=2pt, left=1pt, right=1pt, top=1pt, bottom=1pt, before skip=2pt, after skip=4pt]
\begin{proposition}[Approximate Domain Invariance of Slow Modes]
\label{thm:main_invariance}
Fix a semantic class $c_l$ and let $\xi>0$ be a density threshold, $P_{S,c_l},P_{T,c_l}$ be the source and target class-conditional densities, with high-density regions $\Omega_{S,c_l,\xi}=\{x:P_{S,c_l}(x)\ge\xi\}$ and $\Omega_{T,c_l,\xi}=\{x:P_{T,c_l}(x)\ge\xi\}$, joint core $U_{c_l}=\Omega_{S,c_l,\xi}\cup\Omega_{T,c_l,\xi}$, and $V_{c_l}^{\min}=\min\{|\Omega_{S,c_l,\xi}|,|\Omega_{T,c_l,\xi}|\}$. Let $m_{c_l}=\alpha_{S,c_l}=\alpha_{T,c_l}$ be the balanced class mass, $\Delta C_{c_l}$ the spread of transport costs within the class, and $\ell_{S,c_l}^{\mathrm{OT}},\ell_{T,c_l}^{\mathrm{OT}}<1$ the worst-case source and target transport leakages. Let $f_\lambda\in L^2(\nu)$ be an eigenfunction of $\mathcal{K}$ with eigenvalue $\lambda$. For $V_{c_l}^{\min}>0$, the variance $\mathcal{V}_{U_{c_l}}(f_\lambda)$ of $f_\lambda$ over $U_{c_l}$ satisfies:
\begin{equation}
\mathcal{V}_{U_{c_l}}(f_\lambda)\ \le\ \frac{4(1-\lambda)\,e^{2\Delta C_{c_l}/\varepsilon}}{V_{c_l}^{\min}\,m_{c_l}\,\xi^2\bigl(1-\ell_{S,c_l}^{\mathrm{OT}}\bigr)\bigl(1-\ell_{T,c_l}^{\mathrm{OT}}\bigr)}\,\|f_\lambda\|_{L^2(\nu)}^2 .
\label{eq:main-bound}
\end{equation}
\end{proposition}
\end{tcolorbox}
The bound is obtained in two stages. First, one shows that the affinity $\mathcal{A}(x,y)$ is uniformly lower-bounded on the two high-density regions of the class, by $\tfrac12(\,m_{c_l}\,\xi^2\,e^{-2\Delta C_{c_l}/\varepsilon}(1-\ell_{S,c_l}^{\mathrm{OT}})(1-\ell_{T,c_l}^{\mathrm{OT}}))$, which states that, in the favorable regime, any source and target point of the same class that remain inside their high-density regions stay strongly connected under the dynamics, precisely when the class densities are concentrated, the intra-class transport costs are homogeneous, and the cross-class transport cost is negligible. Second, an eigenfunction whose eigenvalue is close to one carries little Dirichlet energy, so it can only change slowly across the space, and together with the connectivity bound it is forced to stay nearly flat between the source and target cores, which is exactly what bounds its variance over $U_{c_l}$. The bound itself reads intuitively, for a favorable stability regime \textit{s.t.} $1-\lambda \ll V_{c_l}^{\min}m_{c_l}\xi^2e^{-2\Delta C_{c_l}/\varepsilon}(1-\ell_{S,c_l}^{\mathrm{OT}})(1-\ell_{T,c_l}^{\mathrm{OT}})$, a slow mode stays close to constant over the whole class core. Hence, the value taken by such a mode on source and target points belonging to the same high-density class region becomes approximately independent of the domain of origin. This is what lets us read the slow spectral modes as approximately domain-invariant coordinates. We provide the detailed proof of \autoref{thm:main_invariance} in Appendix~\ref{app:theoretical_guarantees_proofs}.
\begin{figure*}[t]
    \centering
    \begin{subfigure}[t]{0.22\textwidth}
        \centering
        \hspace{-0.5cm}
        \includegraphics[width=1.05\textwidth]{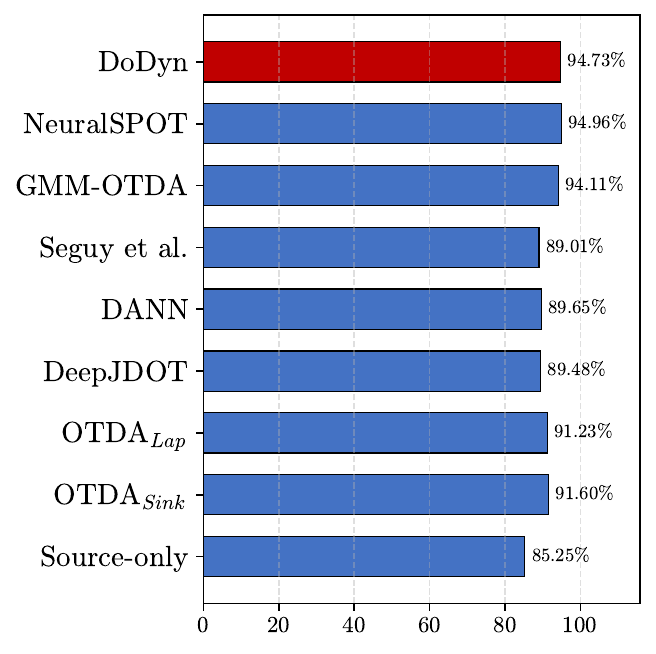}
        \caption{MSD - $\mathcal{A}_{(T)}$}
        \label{fig:msd_at}
    \end{subfigure}
    \hfill
    \begin{subfigure}[t]{0.22\textwidth}
        \centering
        \hspace{-0.5cm}
        \includegraphics[width=1.05\textwidth]{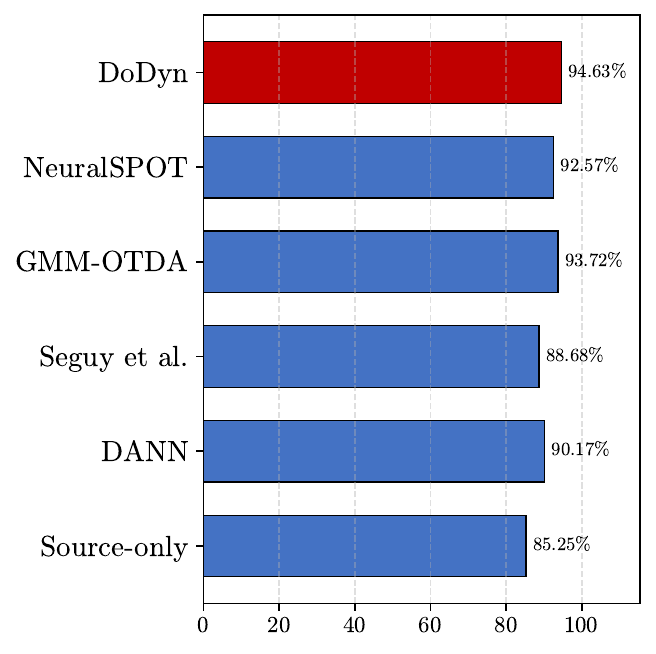}
        \caption{MSD - $\mathcal{A}_{(OOS)}$}
        \label{fig:msd_oos}
    \end{subfigure}
    \hfill
    \begin{subfigure}[t]{0.22\textwidth}
        \centering
        \hspace{-0.5cm}
        \includegraphics[width=1.05\textwidth]{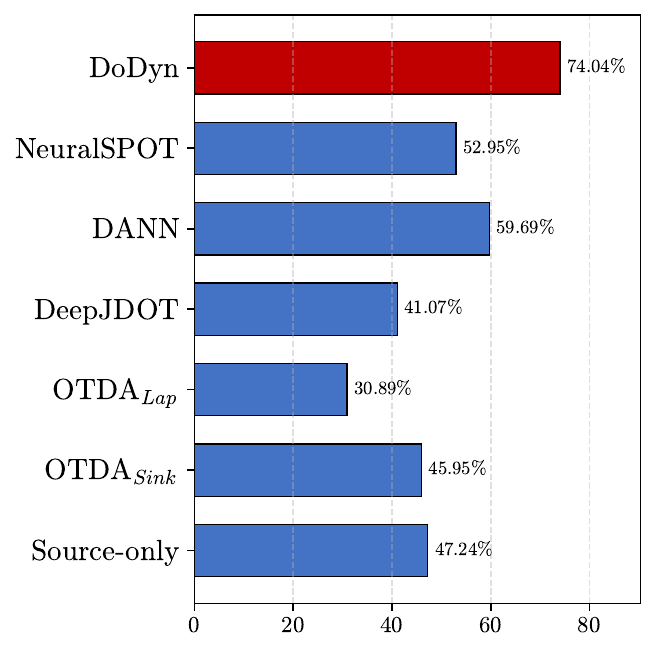}
        \caption{MGR - $\mathcal{A}_{(T)}$}
        \label{fig:mgr_at}
    \end{subfigure}
    \hfill
    \begin{subfigure}[t]{0.22\textwidth}
        \centering
        \hspace{-0.5cm}
        \includegraphics[width=1.05\textwidth]{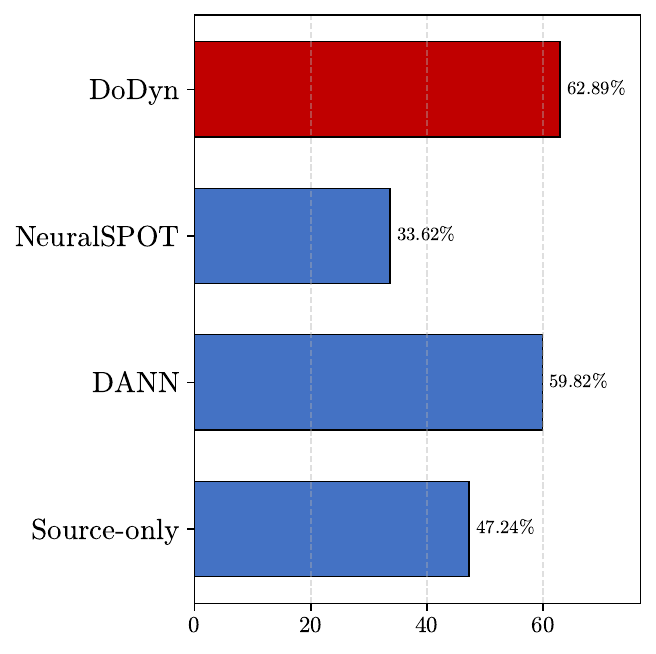}
        \caption{MGR - $\mathcal{A}_{(OOS)}$}
        \label{fig:mgr_oos}
    \end{subfigure}
    \caption{Average adaptation performance (classification accuracy) across all domain pairs on the MSD and MGR benchmarks. Values correspond to the mean accuracy averaged over multiple seeds.}
    \label{fig:avg_results}
\end{figure*}
\subsection{Learning Procedure}\label{sec:learning}
Given source and target samples lifted to Gaussian mixtures and the entropic coupling as described in Section~\ref{sec:emp2gaussian}, we initialize $\boldsymbol{\theta_S} \in \mathbb{R}^{K_S \times k}$ and $\boldsymbol{\theta_T} \in \mathbb{R}^{K_T \times k}$ as learnable parameters, where $k$ is the embeddings dimension. We then estimate the overlap matrices $\mathbf{\widehat{M}_{SS}, \widehat{M}_{ST}, \widehat{M}_{TS}, \widehat{M}_{TT}}$ once via Monte Carlo and keep them fixed throughout training. In addition, to introduce a class discriminative structure, we introduce a classifier $\mathcal{C}_\phi: \mathbb{R}^k 
\to \mathbb{R}^{N_c}$, parameterized as a multi-layer perceptron $\mathcal{C}_\phi = \mathcal{C}_{\phi_L} \circ \cdots \circ 
\mathcal{C}_{\phi_1}$, where $N_c$ is the number of classes. At each iteration, $\hat f_{\boldsymbol{\Theta}}(\mathbf{x}_s^i)$ is computed via \eqref{eq:f_g_form_bis}, all together producing $\hat f_{\boldsymbol{\Theta}}(\mathbf{X_s})$ that are fed forward to 
$\mathcal{C}_\phi$.
Writing $\mathbf{M_\pi}$ for the block operator in \eqref{eq:theta_fP_System_main}, and for a normalized eigenvector $\boldsymbol{\Theta}_{\lambda_\ell}$ associated with the eigenvalue $\lambda_\ell$, one obtains $\|\boldsymbol{\Theta}_{\lambda_\ell}-\mathbf{M_\pi}\boldsymbol{\Theta}_{\lambda_\ell}\|_F^2=(1-\lambda_\ell)^2$, which is small when $\lambda_\ell\lesssim1$. Therefore $\boldsymbol{\Theta}_{\lambda_\ell}$ lies in the ellipsoid $\{\boldsymbol{\Theta}\in \mathbb{R}^{k(K_S+K_T)}, \|\boldsymbol{\Theta}-\mathbf{M_\pi}\boldsymbol{\Theta}\|_F^2\le\epsilon\}$, for some small $\epsilon$.  The classifier here contributes as well by selecting, among functions that lie in this ellipsoid those that also separate classes, producing a classifier-regularized approximation of the slow Koopman embeddings. This motivates the following joint learning optimization problem:
\begin{align}
\underset{\boldsymbol{\Theta},\,\phi}{\operatorname{argmin}}
\quad
\left\{
\begin{aligned}
\|\boldsymbol{\Theta}-\mathbf{M_\pi}\boldsymbol{\Theta}\|_F^2
+\;&\eta\,\mathcal{L}_{\mathrm{CE}}\bigl(\mathcal{C}_\phi(\hat f_{\boldsymbol{\Theta}}(\mathbf{X_s})), \mathbf{Y_s}\bigr)
\end{aligned}\right\}
\label{dodyn_objective}
\end{align}
where the second term is the classification loss, which depends on all the training variables, while the first term acts as a regularizer on the eigenmodes mixing parameters $\boldsymbol{\Theta}$, with a balancing hyperparameter $\eta > 0$.
The full parameter set
$\{\boldsymbol{\theta}_S, \boldsymbol{\theta}_T, \boldsymbol{\phi}\}$ is then optimized jointly by gradient descent \citep{rumelhart1986learning, kingma2014adam}.
This end-to-end training procedure allows the slow modes to be optimized for discriminability while remaining close to the estimated spectral dynamics, thereby accommodating both transport mapping errors and Monte Carlo approximation errors in \(\mathbf{M_\pi}\). At inference, {\textit{no fine-tuning is  required}}. For any new test point $\mathbf{x}$, the corresponding $\hat f(\mathbf{x})$ is reconstructed via \eqref{eq:f_g_form_bis} 
using the learned $\boldsymbol{\{\theta_S, \theta_T\}}$, and classification is performed 
directly via $\mathcal{C}_\phi(\hat f(\mathbf{x}))$.

\section{Experiments}\label{sec:experiments}
In this section, we evaluate \textbf{DoDyn} alongside domain adaptation methods on different UDA tasks.

\textbf{Benchmarks.} We evaluate \textbf{DoDyn} on five benchmarks spanning audio, signals and visual modalities. \textit{Music Genre Recognition (MGR)} \citep{tzanetakis2002musical} and \textit{Music and Speech Detection (MSD)} \citep{Tzanetakis2001AutomaticMG} are audio classification datasets where 
domain shifts arise from five noise conditions. \textit{Compressed Sensing Reflectometry (CS-RT)} \citep{slimani2024detection} is a cable fault diagnosis benchmark based on time domain reflectometry and compressive sensing, where shifts arise from different signal compression factors and cable characteristics. \textit{CMU Pose, Illumination and Expression (CMU-PIE)} \citep{sim2002cmu} is a face recognition benchmark with four camera  viewpoints as domains. Finally, \textit{VisDA} \citep{peng2017visda} is a large-scale synthetic-to-real visual adaptation dataset, pairing 152K synthetic images with 55K real images, which evaluates adaptation at a scale far beyond other benchmarks. Since we are interested in out-of-sample generalization, a portion of each target domain is withheld and used exclusively for OOS evaluation.

\textbf{Baselines.} We compare \textbf{DoDyn} against the following methods: \textbf{Source-only} represents the no adaptation baseline. \textbf{OTDA$_{\text{Sink}}$},  \textbf{OTDA$_{\text{Lap}}$} \citep{courty2017OTDA} , \textbf{DeepJDOT} \citep{damodaran2018deepjdot}, \textbf{DANN} 
\citep{ganin2016domain}, \textbf{Seguy'18}~\citep{seguy2018large} for which we implement their Monge map estimation procedure (Algorithm 1+2 with entropic regularization), \textbf{GMM-OTDA} \cite{montesuma2025optimal}, and \textbf{NeuralSPOT}, a baseline we propose that solves a discrete OT problem and approximates a spectral embedding through a neural network. This comparison covers shallow OT, deep domain adaptation, and neural OT-based methods. Note that not all methods provide out-of-sample extension ($\mathcal{A}_{\text{(OOS)}}$), those that do not are evaluated only on the target domain ($\mathcal{A}_{(T)}$).

\begin{table*}[t]
\centering
\caption{Classification results on the CS-RT benchmark. Each row shows results for a "source $\rightarrow$ target" case, reporting accuracy on the target domain ($\mathcal{A}_{(T)} \uparrow$) and, for the concerned methods, on out-of-sample data ($\mathcal{A}_{(oos)} \uparrow$). Colored values in the average row indicate the performance \textcolor{black}{improvement}/\textcolor{black}{drop} \textit{w.r.t} source-only.}
\small
\setlength{\tabcolsep}{3pt}
\resizebox{\textwidth}{!}{
\begin{tabular}{lccccccccccccccc}
\toprule
\multirow{2}{*}{\textbf{Domains}}
& \multicolumn{1}{c}{\textbf{Source-only}}
& \textbf{OTDA$_{\text{Sink}}$}
& \textbf{OTDA$_{\text{Lap}}$}
& \multicolumn{1}{c}{\textbf{DeepJDOT}}
& \multicolumn{2}{c}{\textbf{DANN}}
& \multicolumn{2}{c}{\textbf{Seguy'18~\cite{seguy2018large}}}
& \multicolumn{2}{c}{\textbf{GMM-OTDA}}
& \multicolumn{2}{c}{\textbf{NeuralSPOT}}
& \multicolumn{2}{c}{\textbf{DoDyn}} \\
\cmidrule(lr){2-2}
\cmidrule(lr){3-3}
\cmidrule(lr){4-4}
\cmidrule(lr){5-5}
\cmidrule(lr){6-7}
\cmidrule(lr){8-9}
\cmidrule(lr){10-11}
\cmidrule(lr){12-13}
\cmidrule(lr){14-15}
& $\mathcal{A}_{(T)} \uparrow$ 
& $\mathcal{A}_{(T)} \uparrow$
& $\mathcal{A}_{(T)} \uparrow$
& $\mathcal{A}_{(T)} \uparrow$
& $\mathcal{A}_{(T)} \uparrow$ & $\mathcal{A}_{(oos)} \uparrow$
& $\mathcal{A}_{(T)} \uparrow$ & $\mathcal{A}_{(oos)} \uparrow$
& $\mathcal{A}_{(T)} \uparrow$ & $\mathcal{A}_{(oos)} \uparrow$
& $\mathcal{A}_{(T)} \uparrow$ & $\mathcal{A}_{(oos)} \uparrow$
& $\mathcal{A}_{(T)} \uparrow$ & $\mathcal{A}_{(oos)} \uparrow$ \\
\midrule

$F_{16} \rightarrow F_{4}$
& $43.60_{\pm 0.42}$ 

& $42.41_{\pm 0.00}$
& $56.10_{\pm 0.84}$
& $43.67_{\pm 2.05}$
& $54.67_{\pm 2.62}$ & $56.33_{\pm 3.27}$
& $28.17_{\pm 1.70}$ & $26.17_{\pm 5.44}$
& \textcolor{black}{$63.67_{\pm 0.62}$}& \textcolor{black}{$56.83_{\pm 0.94}$}
& $72.00_{\pm 0.71}$ & $61.50_{\pm 2.27}$
& $75.50_{\pm 0.41}$ & $79.00_{\pm 0.82}$ \\

$F_{16} \rightarrow F_{2}$
& $43.75_{\pm 0.73}$ 
& $38.54_{\pm 0.56}$
& $54.02_{\pm 0.36}$
& $37.17_{\pm 7.96}$
& $61.33_{\pm 0.85}$ & $62.00_{\pm 1.78}$
& $26.50_{\pm 3.89}$ & $20.67_{\pm 4.09}$
& \textcolor{black}{$60.33_{\pm 0.24}$}& \textcolor{black}{$58.50_{\pm 0.41}$}
& $57.83_{\pm 1.31}$ & $50.67_{\pm 3.66}$
& $65.67_{\pm 0.85}$ & $75.83_{\pm 0.85}$ \\

$F_{16} \rightarrow Ph$
& $48.51_{\pm 2.20}$
& $27.83_{\pm 2.10}$
& $29.46_{\pm 0.36}$
& $39.83_{\pm 3.42}$
& $46.67_{\pm 4.92}$ & $47.00_{\pm 3.63}$
& $25.33_{\pm 4.84}$ & $24.50_{\pm 4.02}$
& \textcolor{black}{$36.67_{\pm 0.24}$}& \textcolor{black}{$47.17_{\pm 4.52}$}
& $24.50_{\pm 0.00}$ & $23.00_{\pm 1.78}$
& $42.00_{\pm 3.08}$ & $57.33_{\pm 5.57}$ \\

\midrule

$F_{4} \rightarrow F_{16}$
& $25.89_{\pm 0.00}$
& $25.45_{\pm 0.00}$
& $68.60_{\pm 0.21}$
& $39.83_{\pm 5.54}$
& $56.83_{\pm 7.19}$ & $58.33_{\pm 3.79}$
& $25.00_{\pm 8.78}$ & $22.00_{\pm 3.94}$
& \textcolor{black}{$46.50_{\pm 1.08}$}& \textcolor{black}{$26.50_{\pm 0.00}$}
& $73.17_{\pm 1.55}$ & $69.17_{\pm 6.64}$
& $73.50_{\pm 0.41}$ & $76.00_{\pm 1.63}$ \\

$F_{4} \rightarrow F_{2}$
& $29.76_{\pm 0.21}$
& $28.72_{\pm 0.76}$
& $30.80_{\pm 6.65}$
& $39.17_{\pm 4.87}$
& $56.33_{\pm 3.79}$ & $52.67_{\pm 3.12}$
& $17.67_{\pm 2.09}$ & $24.83_{\pm 10.07}$
& \textcolor{black}{$55.33_{\pm 1.18}$}& \textcolor{black}{$48.67_{\pm 0.62}$}
& $52.00_{\pm 4.08}$ & $55.33_{\pm 6.33}$
& $59.50_{\pm 0.71}$ & $63.17_{\pm 1.31}$ \\

$F_{4} \rightarrow Ph$
& $36.61_{\pm 0.73}$
& $18.75_{\pm 0.00}$
& $30.36_{\pm 0.36}$
& $32.50_{\pm 2.68}$
& $33.17_{\pm 3.27}$ & $34.67_{\pm 0.47}$
& $27.00_{\pm 8.20}$ & $25.33_{\pm 8.59}$
& \textcolor{black}{$38.83_{\pm 1.55}$}& \textcolor{black}{$40.67_{\pm 0.62}$}
& $31.67_{\pm 6.33}$ & $32.83_{\pm 7.15}$
& $38.83_{\pm 5.72}$ & $54.50_{\pm 3.34}$ \\

\midrule

$F_{2} \rightarrow F_{16}$
& $26.79_{\pm 0.00}$ 
& $40.18_{\pm 0.00}$
& $27.98_{\pm 0.76}$
& $37.00_{\pm 8.15}$
& $54.50_{\pm 7.76}$ & $54.33_{\pm 6.79}$
& $23.00_{\pm 6.48}$ & $30.17_{\pm 3.52}$
& \textcolor{black}{$34.83_{\pm 0.47}$}& \textcolor{black}{$26.00_{\pm 0.00}$}
& $58.67_{\pm 0.62}$ & $61.17_{\pm 7.00}$
& $63.17_{\pm 0.94}$ & $67.00_{\pm 0.00}$ \\

$F_{2} \rightarrow F_{4}$
& $27.38_{\pm 0.21}$
& $25.45_{\pm 0.00}$
& $37.05_{\pm 0.00}$
& $36.00_{\pm 7.22}$
& $52.17_{\pm 6.51}$ & $48.67_{\pm 6.51}$
& $30.00_{\pm 4.71}$ & $31.50_{\pm 4.71}$
& \textcolor{black}{$42.17_{\pm 0.24}$}& \textcolor{black}{$45.33_{\pm 0.62}$}
& $54.17_{\pm 1.31}$ & $60.33_{\pm 1.84}$
& $56.33_{\pm 0.94}$ & $64.50_{\pm 1.08}$ \\

$F_{2} \rightarrow Ph$
& $66.52_{\pm 1.89}$ 
& $38.10_{\pm 0.84}$
& $72.02_{\pm 0.42}$
& $57.50_{\pm 9.90}$
& $76.33_{\pm 5.39}$ & $82.83_{\pm 1.93}$
& $62.67_{\pm 3.57}$ & $54.33_{\pm 5.91}$
& \textcolor{black}{$57.00_{\pm 0.82}$} & \textcolor{black}{$62.00_{\pm 0.71}$}
& $60.33_{\pm 3.70}$ & $55.33_{\pm 6.33}$
& $61.83_{\pm 6.25}$ & $77.33_{\pm 2.90}$ \\

\midrule

$Ph \rightarrow F_{16}$
& $26.34_{\pm 0.00}$ 
& $21.13_{\pm 0.42}$
& $30.06_{\pm 0.21}$
& $36.67_{\pm 7.77}$
& $46.50_{\pm 5.10}$ & $45.00_{\pm 3.19}$
& $29.17_{\pm 2.39}$ & $27.33_{\pm 10.27}$
& \textcolor{black}{$33.83_{\pm 0.24}$}& \textcolor{black}{$25.67_{\pm 0.24}$}
& $34.50_{\pm 4.26}$ & $29.17_{\pm 7.58}$
& $55.83_{\pm 1.55}$ & $48.67_{\pm 3.06}$ \\

$Ph \rightarrow F_{4}$
& $26.34_{\pm 0.00}$
& $20.98_{\pm 0.34}$
& $43.30_{\pm 0.96}$
& $37.17_{\pm 1.65}$
& $40.17_{\pm 3.06}$ & $38.33_{\pm 2.46}$
& $23.00_{\pm 3.34}$ & $22.17_{\pm 8.25}$
& \textcolor{black}{$38.00_{\pm 0.71}$}& \textcolor{black}{$43.67_{\pm 0.24}$}
& $26.83_{\pm 9.57}$ & $30.67_{\pm 10.07}$
& $43.00_{\pm 0.71}$ & $46.67_{\pm 2.25}$ \\

$Ph \rightarrow F_{2}$
& $46.73_{\pm 0.76}$ 
& $50.74_{\pm 0.42}$
& $61.31_{\pm 0.92}$
& $48.50_{\pm 7.78}$
& $63.17_{\pm 2.01}$ & $58.50_{\pm 0.41}$
& $50.67_{\pm 11.83}$ & $39.83_{\pm 13.12}$
& \textcolor{black}{$62.83_{\pm 0.47}$}& \textcolor{black}{$56.50_{\pm 0.82}$}
& $61.50_{\pm 2.12}$ & $51.17_{\pm 3.70}$
& $66.67_{\pm 2.09}$ & $66.50_{\pm 1.87}$ \\

\midrule
\rowcolor{gray!10}\textbf{Avg.}
& $37.35$ 
& $33.11$ \textcolor{red}{($4.24\downarrow$)}
& $45.09$ \textcolor{darkgreen}{($7.74\uparrow$)}
& $40.42$ \textcolor{darkgreen}{($3.07\uparrow$)}
& $\underline{53.49}$ \textcolor{darkgreen}{($16.14\uparrow$)} & $53.22$ \textcolor{darkgreen}{($15.56\uparrow$)}
& $30.68$ \textcolor{red}{($6.67\downarrow$)} & $29.07$ \textcolor{red}{($8.59\downarrow$)}
& \textcolor{black}{$47.50$} \textcolor{darkgreen}{($10.15 \uparrow$)}& \textcolor{black}{$44.79$} \textcolor{darkgreen}{($7.44\uparrow$)}
& $50.60$ \textcolor{darkgreen}{($12.26\uparrow$)} & $48.36$ \textcolor{darkgreen}{($10.70\uparrow$)}
& $\mathbf{58.49}$ \textcolor{darkgreen}{($21.14\uparrow$)} & $\mathbf{64.71}$ \textcolor{darkgreen}{($27.05\uparrow$)} \\
\bottomrule
\end{tabular}
}
\label{tab:results_CSRT}
\end{table*}
\begin{wraptable}{r}{0.5\textwidth}
\centering
\caption{Classification results on the CMU-PIE benchmark, averaged over multiple seeds.}
\small
\setlength{\tabcolsep}{0.0001pt}
\resizebox{\linewidth}{!}{
\begin{tabular}{lcccccccc}
\toprule
\multirow{2}{*}{\textbf{Domains}}
& \textbf{Source-only}
& \textbf{OTDA$_{\text{Sink}}$}
& \multicolumn{2}{c}{\textbf{DANN}}
& \multicolumn{2}{c}{\textcolor{black}{\textbf{GMM-OTDA}}}
& \multicolumn{2}{c}{\textbf{DoDyn}} \\
\cmidrule(lr){2-2}
\cmidrule(lr){3-3}
\cmidrule(lr){4-5}
\cmidrule(lr){6-7}
\cmidrule(lr){8-9}

& $\mathcal{A}_{(T)} \uparrow$
& $\mathcal{A}_{(T)} \uparrow$
& $\mathcal{A}_{(T)} \uparrow$ 
&$\mathcal{A}_{(oos)} \uparrow$
& $\mathcal{A}_{(T)} \uparrow$ 
&$\mathcal{A}_{(oos)} \uparrow$
& $\mathcal{A}_{(T)} \uparrow$ 
& $\mathcal{A}_{(oos)} \uparrow$ \\

\midrule
$\text{c07} \rightarrow \text{c29}$
& $25.83$
& $36.40_{\pm 0.79}$
& $26.32_{\pm 3.23}$ & $24.20_{\pm 2.12}$
& \textcolor{black}{$61.21_{\pm 1.10}$} & \textcolor{black}{$55.88_{\pm 2.29}$}
& ${57.26_{\pm 0.87}}$ & ${58.96_{\pm 0.92}}$ \\
$\text{c07} \rightarrow \text{c05}$
& $34.10$
& $34.41_{\pm 1.93}$
& $25.45_{\pm 2.21}$ & $26.47_{\pm 2.19}$
& \textcolor{black}{$34.81_{\pm 0.29}$} & \textcolor{black}{$28.14_{\pm 3.36}$}
& ${44.00_{\pm 1.29}}$ & ${39.31_{\pm 1.91}}$ \\
$\text{c07} \rightarrow \text{c09}$
& $37.41$
& $40.75_{\pm 1.17}$
& $34.56_{\pm 1.28}$ & $32.17_{\pm 0.79}$
& \textcolor{black}{$72.36_{\pm 1.91}$} & \textcolor{black}{$65.59_{\pm 1.10}$}
& ${67.86_{\pm 0.97}}$ & ${70.24_{\pm 0.78}}$ \\
\midrule
$\text{c29} \rightarrow \text{c07}$
& $22.62$
& $14.24_{\pm 3.21}$
& $36.37_{\pm 1.24}$ & $36.96_{\pm 1.94}$
& \textcolor{black}{$59.87_{\pm 0.72}$} & \textcolor{black}{$53.92_{\pm 2.80}$}
& ${58.50_{\pm 0.64}}$ & ${58.87_{\pm 1.11}}$ \\
$\text{c29} \rightarrow \text{c05}$
& $27.02$
& $27.85_{\pm 1.01}$
& $28.84_{\pm 0.12}$ & $27.97_{\pm 0.70}$
& \textcolor{black}{$26.21_{\pm 0.99}$} & \textcolor{black}{$20.98_{\pm 2.07}$}
& ${31.25_{\pm 1.02}}$ & ${27.88_{\pm 2.15}}$ \\
$\text{c29} \rightarrow \text{c09}$
& $29.60$
& $38.20_{\pm 0.87}$
& $30.82_{\pm 2.95}$ & $30.09_{\pm 2.00}$
& \textcolor{black}{$59.75_{\pm 0.99}$} & \textcolor{black}{$57.55_{\pm 2.57}$}
& ${59.74_{\pm 0.26}}$ & ${59.74_{\pm 1.43}}$ \\
\midrule
$\text{c05} \rightarrow \text{c07}$
& $24.82$
& $14.78_{\pm 4.51}$
& $37.08_{\pm 2.05}$ & $35.73_{\pm 0.91}$
& \textcolor{black}{$38.43_{\pm 0.86}$} & \textcolor{black}{$33.24_{\pm 3.00}$}
& ${43.12_{\pm 0.77}}$ & ${44.63_{\pm 1.74}}$ \\
$\text{c05} \rightarrow \text{c29}$
& $19.30$
& $10.88_{\pm 2.45}$
& $29.69_{\pm 0.20}$ & $28.49_{\pm 2.33}$
& \textcolor{black}{$30.26_{\pm 0.71}$} & \textcolor{black}{$26.86_{\pm 1.70}$}
& ${31.83_{\pm 0.34}}$ & ${33.70_{\pm 1.74}}$ \\
$\text{c05} \rightarrow \text{c09}$
& $20.68$
& $9.38_{\pm 2.48}$
& $27.94_{\pm 4.37}$ & $30.09_{\pm 3.69}$
& \textcolor{black}{$37.93_{\pm 1.37}$} & \textcolor{black}{$33.53_{\pm 2.29}$}
& ${44.49_{\pm 0.78}}$ & ${46.69_{\pm 0.78}}$ \\
\midrule
$\text{c09} \rightarrow \text{c07}$
& $28.23$
& $18.58_{\pm 0.69}$
& $34.81_{\pm 0.92}$ & $34.25_{\pm 0.60}$
& \textcolor{black}{$71.01_{\pm 0.96}$} & \textcolor{black}{$59.80_{\pm 3.40}$}
& ${67.03_{\pm 0.99}}$ & ${67.40_{\pm 1.30}}$ \\
$\text{c09} \rightarrow \text{c29}$
& $33.09$
& $35.72_{\pm 1.06}$
& $30.12_{\pm 1.81}$ & $30.76_{\pm 3.70}$
& \textcolor{black}{$62.56_{\pm 0.54}$} & \textcolor{black}{$57.06_{\pm 3.23}$}
& ${59.50_{\pm 0.49}}$ & ${60.42_{\pm 0.96}}$ \\
$\text{c09} \rightarrow \text{c05}$
& $32.26$
& $30.06_{\pm 0.57}$
& $27.96_{\pm 1.74}$ & $29.05_{\pm 1.39}$
& \textcolor{black}{$33.64_{\pm 1.04}$} & \textcolor{black}{$57.06_{\pm 3.23}$}
& ${42.13_{\pm 0.53}}$ & ${37.56_{\pm 0.09}}$ \\
\midrule
\rowcolor{gray!10}\textbf{Avg.}
& $27.91$
& $25.94\;\textcolor{red}{(1.97\downarrow)}$ \;
& ${30.83}\;\textcolor{darkgreen}{(2.92\uparrow)}$ \; & ${30.52}\;\textcolor{darkgreen}{(2.61\uparrow)} \;$
& $\underline{49.00}\;\textcolor{darkgreen}{(21.09\uparrow)} \;$
& $\underline{43.08}\;\textcolor{darkgreen}{(15.17\uparrow)} \;$
& 
$\mathbf{50.56}\;\textcolor{darkgreen}{(22.65\uparrow)}\;$ & $\mathbf{50.45}\;\textcolor{darkgreen}{(22.54\uparrow)}$ \\
\bottomrule
\end{tabular}
}
\label{tab:results_CMUc}
\end{wraptable}
\textbf{Main Results.} For MGR, MSD, CSRT, and CMU-PIE, we use each sample as a gaussian mixture component's mean (which implies $K_S=n_s$ and $K_T=n_t$ in this setup), {with isotropic covariance matrixe $\sigma^2 I$}. Figure~\ref{fig:avg_results} reports average adaptation performance on MSD and MGR. On MSD (Fig.~\ref{fig:msd_at} and~\ref{fig:msd_oos}), all methods improve over source-only, shallow OT methods, DeepJDOT, \citep{seguy2018large} and DANN  perform reasonably well alongside GMM-OTDA, NeuralSPOT and DoDyn which achieves the best (and comparable) performance, with DoDyn's advantage on out-of-sample generalization. MGR (Fig.~\ref{fig:mgr_at} and~\ref{fig:mgr_oos}) is a harder setting, since transport plan becomes more sensitive to mismatches as class complexity grows \citep{montesuma2021wasserstein, sadsaoud2026beyond}. Shallow OT methods and DeepJDOT now fall slightly below source-only, DANN remains competitive. Here the gap compared to DoDyn increases, especially on out-of-sample evaluation. On CS-RT benchmark, Table~\ref{tab:results_CSRT} shows that OTDA$_{\text{sink}}$ and \citep{seguy2018large} are below source-only, while OTDA$_{\text{Lap}}$, DeepJDOT, DANN, GMM-OTDA and  NeuralSPOT are improving, while DoDyn gets the best performance on both and DANN achieves the second best performance. On CMU-PIE (Table~\ref{tab:results_CMUc}) OTDA$_{\text{Sink}}$ under-performs source-only and DANN gains are modest. Here, GMM-OTDA achieves the second best average performance while DoDyn leads on target and OOS. Finally, since VisDA is a large scale benchmark, we run our algorithm using GMMs with 48 component on each domain (under the same setup of \cite{montesuma2025optimal}). Fig.\ref{fig:visda_results} shows that DANN, GMM-OTDA and DoDyn improve the adaptation performance by nearly $15\%$ and achieve a comparable performance on both target and out-of-sample settings. 
\begin{figure*}[t]
    \centering
    \begin{subfigure}[h]{0.25\textwidth}
        \centering
        \includegraphics[width=\textwidth]{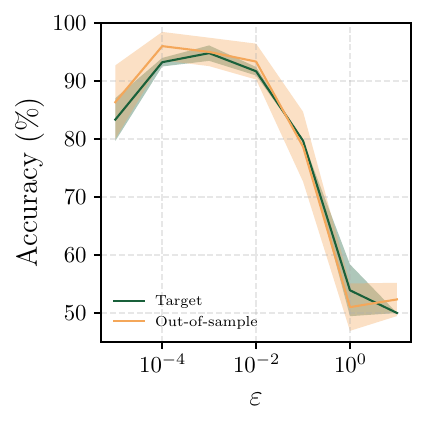}
        \label{fig:ablation_eps}
    \end{subfigure}
    \hfill
    \begin{subfigure}[h]{0.25\textwidth}
        \centering
        \includegraphics[width=\textwidth]{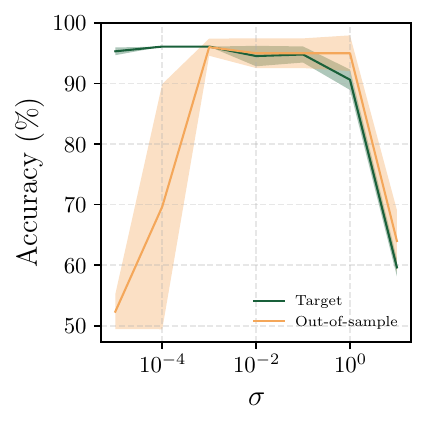}
        \label{fig:ablation_sigma}
    \end{subfigure}
    \hfill
    \begin{subfigure}[h]{0.23\textwidth}
    \vspace{-0.1cm}
    \includegraphics[width=\textwidth]{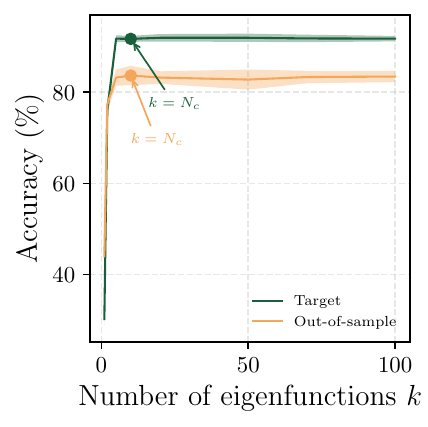}
            \label{fig:ablation_number_eigenfucntions_k}
        \end{subfigure}
    \vspace{-0.5cm}
    \caption{Sensitivity of \textbf{DoDyn} to the entropic regularization parameter $\varepsilon$, the standard deviation $\sigma$, and the number of eigenfunctions $k$.}
    \label{fig:ablation}
\end{figure*}

\textbf{Ablation.} Figure~\ref{fig:ablation} illustrates DoDyn's sensitivity to its main hyperparameters and design choices on the MSD benchmark. Varying the entropic regularization parameter $\varepsilon$ over $\{10^{-5}, \dots, 10\}$ (Fig.~\ref{fig:ablation}a) illustrates a stable, near-optimal accuracy for $\varepsilon \in [10^{-4}, 10^{-2}]$ on both target and OOS evaluation, while as $\varepsilon \rightarrow 0$, the entropic OT converges to exact OT, where each source point is  mapped
\begin{wrapfigure}[12]{r}{0.5\textwidth}
  \centering
  \vspace{-0.3cm}
  \begin{subfigure}[t]{0.24\textwidth}
    \includegraphics[width=\textwidth]{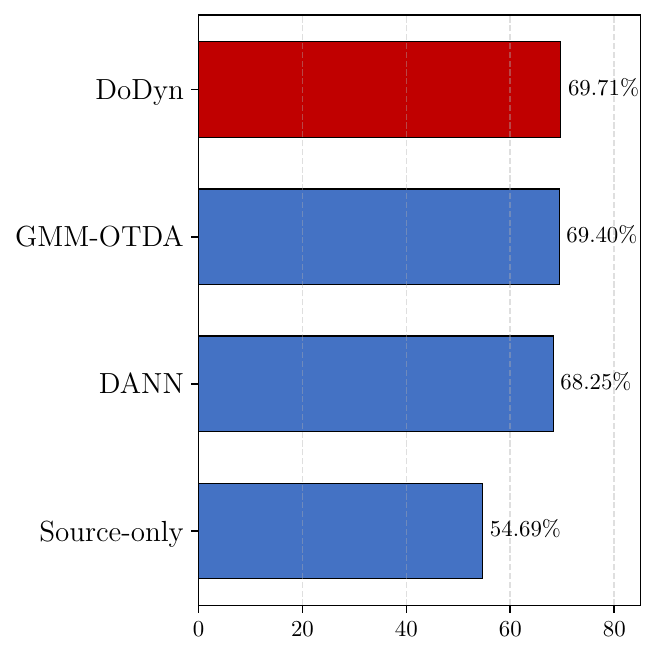}
  \end{subfigure}
  \hfill
  \begin{subfigure}[t]{0.24\textwidth}
    \includegraphics[width=\textwidth]{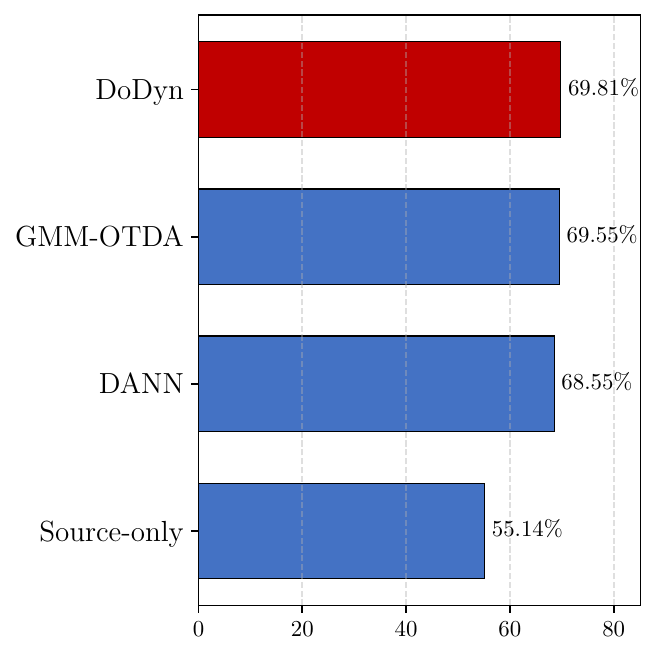}
  \end{subfigure}
  \caption{Adaptation performance on VisDA benchmark.}
  \label{fig:visda_results}
\end{wrapfigure}
 to a single target point. In graph terms, this gives isolated point pairs as connected components. At the other extreme, large $\varepsilon$ enforces a near-uniform mass distribution across all points, collapsing the entire source-target graph into a single  component, the resulting embedding carries no discriminative structure, and performance drops sharply. In the setup of isotropic lifting, varying $\sigma$ (Fig.~\ref{fig:ablation}b) follows a similar pattern, stable for $\sigma \in [10^{-4}, 1]$ and degrading at both extremes, where small $\sigma$ collapses the mixtures into Diracs, and the reconstruction weights $g_{Sk_1}, g_{Tk_2}$ in~\eqref{eq:f_g_form_bis} become ill-defined for unseen points, while large $\sigma$ causes excessive overlap between components and washes out those weights. Finally, we evaluate the effect of the number of eigenfunctions $k$ (embedding dimensions). Ideally, one would set $k = N_c$ where $N_c$ is the number of classes, as we are looking for $N_c$ metastable sets \citep{schutte2005uncoupling, Huisinga2006}. We consider the MGR dataset for the domain pair (F16 $\rightarrow$ D), we vary $k\in\{1,2,5,10,20,50,100\}$ and report $\mathcal{A}_T$ and $\mathcal{A}_{\text{oos}}$ in Fig.~\ref{fig:ablation}c. As shown, performance is poor for $k < N_c = 10$. As $k$ grows, performance starts improving, reaching its best at $k = N_c$. Beyond this point, performance plateaus and remains stable regardless of how many additional eigenfunctions are computed. The relevant spectral information is captured by the first $N_c$ eigenfunctions, and adding more does not degrade nor improve the representation. This also gives a simple and intuitive rule where $k$ is simply set to the number of classes without any tuning.

\section{Conclusion}\label{sec:conclusion}
We introduced DoDyn, a domain adaptation method that learns cross-domain representations as closed-form eigenfunctions of a Koopman operator defined on a random walk induced by an optimal transport coupled graph, and allows accurate out-of-sample embedding without recomputing the transport plan at inference. Empirical validation on several benchmarks for acoustic, real-world signal processing and visual adaptation tasks shows DoDyn achieves a good performance compared to shallow OT, deep domain adaptation and neural OT baselines, with an out-of-sample generalization.

\section*{Acknowledgments and Disclosure of Funding}
CGP has benefited from French State aid managed by the Agence Nationale de la Recherche (ANR) under France 2030 program with the reference ANR-23-PEIA-005 (REDEEM project).
\bibliography{main}
\bibliographystyle{plain}
\newpage
\appendix
\section{Proofs of Theoretical Results}
\subsection{Proof of Lemma~\ref{lem:marginals}}\label{App:proof_marginals}
The marginals of $\pi_{\varepsilon}(x_s, x_t)$ are obtained by integrating 
Eq.~\ref{continuous_coupling} with respect to $x_t$, and using that 
each $P_{Tk_2}$ integrates to one:
\begin{subequations}\label{eq:marginal_proof}
\begin{align}
\int_{\mathbb{R}^d}\pi_{\varepsilon}(x_s, x_t)\,dx_t
&=
\sum_{k_1=1}^{K_S}\sum_{k_2=1}^{K_T}\pi_{k_1,k_2}\,P_{Sk_1}(x_s)
\int_{\mathbb{R}^d}P_{Tk_2}(x_t)\,dx_t \\
&=
\sum_{k_1=1}^{K_S}\sum_{k_2=1}^{K_T}\pi_{k_1,k_2}\,P_{Sk_1}(x_s) \\
&=
\underbrace{\sum_{k_1=1}^{K_S}\alpha_{sk_1}\,P_{Sk_1}(x_s)}_{P_S(x_s)},
\end{align}
\end{subequations}
where the last equality uses the marginal constraint 
$\sum_{k_2=1}^{K_T}\pi_{k_1,k_2} = \alpha_{sk_1}$. The analogous 
computation for $\int_{\mathbb{R}^d}\pi_{\varepsilon}(x_t, x_s)\,dx_t$ yields $P_T(x_s)$.

\subsection{Proof of Proposition~\ref{propo:expansion finite}}
\label{App:proof_expansion finite}

    Let the component expectations be defined as: 
    \begin{equation}\label{theta}
    \theta_{S k_1}=\int_{\mathbb{R}^d}f_{{\ell}}(y)\,P_{S k_1}(y)\,dy,
    \qquad
    \theta_{T k_2}=\int_{\mathbb{R}^d}f_{{\ell}}(y)\,P_{T k_2}(y)\,dy.
    \end{equation}
    Replacing the first term integral in Eq.\ref{koopman_b} using Eqs.\ref{continuous_coupling} and \ref{theta} gives:
    \begin{subequations}\label{pi_x_expansion}
    \begin{align}
    \int_{\mathbb{R}^d}\pi(x,y)f_{\ell}(y)\,dy
    &=
    \int
    \Bigg[
    \sum_{k_1=1}^{K_S}\sum_{k_2=1}^{K_T}
    \pi_{k_1,k_2}\,P_{S k_1}(x)\,P_{T k_2}(y)
    \Bigg]f_{\ell}(y)\,dy
    \label{pi_x_expansion_a}\\
    &=
    \sum_{k_1=1}^{K_S}\sum_{k_2=1}^{K_T}
    \pi_{k_1,k_2}\,P_{S k_1}(x)
    \int_{\mathbb{R}^d}P_{T k_2}(y)f_{\ell}(y)\,dy
    \label{pi_x_expansion_b}\\
    &=
    \sum_{k_1=1}^{K_S}\sum_{k_2=1}^{K_T}
    \pi_{k_1,k_2}\,P_{S k_1}(x)\,\theta_{T k_2}.
    \label{pi_x_expansion_c}
    \end{align}
    \end{subequations}
    Similarly for the second integral term,
    \begin{subequations}\label{pi_y_expansion}
    \begin{align}
    \int_{\mathbb{R}^d}\pi(y,x)f_{\ell}(y)\,dy
    &=
    \int
    \Bigg[
    \sum_{k_1=1}^{K_S}\sum_{k_2=1}^{K_T}
    \pi_{k_1,k_2}\,P_{S k_1}(y)\,P_{T k_2}(x)
    \Bigg]f_{\ell}(y)\,dy
    \label{pi_y_expansion_a}\\
    &=
    \sum_{k_1=1}^{K_S}\sum_{k_2=1}^{K_T}
    \pi_{k_1,k_2}\,P_{T k_2}(x)
    \int_{\mathbb{R}^d}P_{S k_1}(y)f_{\ell}(y)\,dy
    \label{pi_y_expansion_b}\\
    &=
    \sum_{k_1=1}^{K_S}\sum_{k_2=1}^{K_T}
    \pi_{k_1,k_2}\,P_{T k_2}(x)\,\theta_{S k_1}.
    \label{pi_y_expansion_c}
    \end{align}
    \end{subequations}
    Plugging both expansions \eqref{pi_x_expansion_c}, \eqref{pi_y_expansion_c} into Eq.\ref{koopman_b}, we obtain:
    \begin{equation}
    \lambda_\ell f_{\ell}(x)=\frac{1}{P_S(x)+P_T(x)}
    \sum_{k_1=1}^{K_S}\sum_{k_2=1}^{K_T}
    \pi_{k_1,k_2}\,P_{S k_1}(x)\,\theta_{T k_2}
    +
    \sum_{k_1=1}^{K_S}\sum_{k_2=1}^{K_T}
    \pi_{k_1,k_2}\,P_{T k_2}(x)\,\theta_{S k_1}.
    \label{eq:f_closed_form_intermediate}
    \end{equation}
    Moreover, it is convenient to define 
    \begin{equation}
    g_{S k_1}(x)=\frac{P_{S k_1}(x)}{P_S(x)+P_T(x)},
    \qquad
    g_{T k_2}(x)=\frac{P_{T k_2}(x)}{P_S(x)+P_T(x)}.
    \label{eq:g_basis}
    \end{equation}
    Then Eq.\ref{eq:f_closed_form_intermediate} rewrites as:
    \begin{equation}
    \lambda_\ell f_{\ell}(x)=
    \left[\sum_{k_1=1}^{K_S} g_{S k_1}(x)\,\sum_{k_2=1}^{K_T}\pi_{k_1,k_2}\theta_{T k_2}
    +
    \sum_{k_2=1}^{K_T} g_{T k_2}(x)\,\sum_{k_1=1}^{K_S}\pi_{k_1,k_2}\theta_{S k_1}\right].
    \label{eq:f_g_closed_form}
\end{equation}

\medskip

\subsection{Derivation of Fixed-Point System~\ref{eq:theta_fP_System_main}}
\label{App:theta_derivation}

\textbf{Source component expectations.} For a fixed $i \in \{1, 
\dots, K_S\}$, by definition:
\begin{equation}
\theta_{Si} = \int_{\mathbb{R}^d} P_{Si}(x)\,f_{\ell}(x)\,dx.
\end{equation}
Substituting the fixed-point expansion of Eq.~\ref{eq:f_g_closed_form} 
into the above gives:
\begin{equation}\label{eq:theta_Si}
\resizebox{0.92\linewidth}{!}{$\displaystyle
\begin{aligned}
\theta_{Si}
&= \int_{\mathbb{R}^d} P_{Si}(x)\!\left[\frac{1}{\lambda_\ell}\sum_{k_1=1}^{K_S} g_{Sk_1}(x)\sum_{k_2=1}^{K_T}\pi_{k_1,k_2}\theta_{Tk_2}
+ \frac{1}{\lambda_\ell}\sum_{k_2=1}^{K_T} g_{Tk_2}(x)\sum_{k_1=1}^{K_S}\pi_{k_1,k_2}\theta_{Sk_1}\right]dx \\
&= \frac{1}{\lambda_\ell}\!\left[\sum_{k_1=1}^{K_S}\sum_{k_2=1}^{K_T}\!\left(\int_{\mathbb{R}^d} P_{Si}(x)\,g_{Sk_1}(x)\,dx\right)\pi_{k_1,k_2}\theta_{Tk_2}
+ \sum_{k_2=1}^{K_T}\sum_{k_1=1}^{K_S}\!\left(\int_{\mathbb{R}^d} P_{Si}(x)\,g_{Tk_2}(x)\,dx\right)\pi_{k_1,k_2}\theta_{Sk_1}\right]
\end{aligned}
$}
\end{equation}
where we exchanged the integral and the sums by linearity of 
integration, and factored out $\frac{1}{\lambda_\ell}$. Expanding $g_{Sk_1}(x)$ and $g_{Tk_2}(x)$ using their 
definitions~\eqref{eq:g_basis}, the integrals in \ref{eq:theta_Si} become:
\begin{equation}
\int_{\mathbb{R}^d} P_{Si}(x)\,g_{Sk_1}(x)\,dx 
= \int_{\mathbb{R}^d} \frac{P_{Si}(x)\,P_{Sk_1}(x)}{P_S(x)+P_T(x)}\,dx
= M_{SS}[i,k_1],
\end{equation}
\begin{equation}
\int_{\mathbb{R}^d} P_{Si}(x)\,g_{Tk_2}(x)\,dx 
= \int_{\mathbb{R}^d} \frac{P_{Si}(x)\,P_{Tk_2}(x)}{P_S(x)+P_T(x)}\,dx
= M_{ST}[i,k_2],
\end{equation}
which are exactly the overlap matrices defined in~\eqref{eq:MSS_MST}. 
Substituting them back yields:
\begin{equation}
\lambda_\ell\,\theta_{Si} = \sum_{k_1=1}^{K_S} M_{SS}[i,k_1]
\sum_{k_2=1}^{K_T}\pi_{k_1,k_2}\theta_{Tk_2}
+ \sum_{k_2=1}^{K_T} M_{ST}[i,k_2]
\sum_{k_1=1}^{K_S}\pi_{k_1,k_2}\theta_{Sk_1}.
\label{final_theta_Si}
\end{equation}

\textbf{Target component expectations.} Similarly, for a fixed $j \in \{1, 
\dots, K_T\}$, by definition:
\begin{equation}
\theta_{Tj} = \int_{\mathbb{R}^d} P_{Tj}(x)\,f_{\ell}(x)\,dx.
\end{equation}
Substituting Eq.~\ref{eq:f_g_closed_form} and exchanging integral and 
sums by linearity gives:
\begin{equation}\label{eq:theta_Tj}
\resizebox{0.92\linewidth}{!}{$\displaystyle
\begin{aligned}
\theta_{Tj}
&= \int_{\mathbb{R}^d} P_{Tj}(x)\!\left[\frac{1}{\lambda_\ell}\sum_{k_1=1}^{K_S} g_{Sk_1}(x)\sum_{k_2=1}^{K_T}\pi_{k_1,k_2}\theta_{Tk_2}
+ \frac{1}{\lambda_\ell}\sum_{k_2=1}^{K_T} g_{Tk_2}(x)\sum_{k_1=1}^{K_S}\pi_{k_1,k_2}\theta_{Sk_1}\right]dx \\
&= \frac{1}{\lambda_\ell}\!\left[\sum_{k_1=1}^{K_S}\sum_{k_2=1}^{K_T}\!\left(\int_{\mathbb{R}^d} P_{Tj}(x)\,g_{Sk_1}(x)\,dx\right)\pi_{k_1,k_2}\theta_{Tk_2}
+ \sum_{k_2=1}^{K_T}\sum_{k_1=1}^{K_S}\!\left(\int_{\mathbb{R}^d} P_{Tj}(x)\,g_{Tk_2}(x)\,dx\right)\pi_{k_1,k_2}\theta_{Sk_1}\right]
\end{aligned}
$}
\end{equation}
Expanding $g_{Sk_1}(x)$ and $g_{Tk_2}(x)$ using~\eqref{eq:g_basis} yields:
\begin{equation}
\int_{\mathbb{R}^d} P_{Tj}(x)\,g_{Sk_1}(x)\,dx 
= \int_{\mathbb{R}^d} \frac{P_{Tj}(x)\,P_{Sk_1}(x)}{P_S(x)+P_T(x)}\,dx
= M_{TS}[j,k_1],
\end{equation}
\begin{equation}
\int_{\mathbb{R}^d} P_{Tj}(x)\,g_{Tk_2}(x)\,dx 
= \int_{\mathbb{R}^d} \frac{P_{Tj}(x)\,P_{Tk_2}(x)}{P_S(x)+P_T(x)}\,dx
= M_{TT}[j,k_2],
\end{equation}
which are exactly the overlap matrices defined in~\eqref{eq:MTS_MTT}. 
Substituting back and multiplying both sides by $\lambda_\ell$ yields:
\begin{equation}
\lambda_\ell\,\theta_{Tj} = \sum_{k_1=1}^{K_S} M_{TS}[j,k_1]
\sum_{k_2=1}^{K_T}\pi_{k_1,k_2}\theta_{Tk_2}
+ \sum_{k_2=1}^{K_T} M_{TT}[j,k_2]
\sum_{k_1=1}^{K_S}\pi_{k_1,k_2}\theta_{Sk_1}.
\label{final_theta_Tj}
\end{equation}

Collecting Eqs.~\ref{final_theta_Si}--\ref{final_theta_Tj} in vector 
form gives:
\begin{equation}
\begin{aligned}
\lambda_\ell\,\boldsymbol{\theta_S} &= \mathbf{M}_{SS}\,\boldsymbol{\pi}\,\boldsymbol{\theta}_T +\mathbf{M}_{ST}\,\boldsymbol{\pi}^\top\boldsymbol{\theta}_S,\\
\lambda_\ell\,\boldsymbol{\theta_T} &= \mathbf{M}_{TS}\,\boldsymbol{\pi}\,\boldsymbol{\theta}_T 
+ \mathbf{M}_{TT}\,\boldsymbol{\pi}^\top\boldsymbol{\theta}_S,
\end{aligned}
\quad \boldsymbol{\Theta} = 
[\boldsymbol{\theta}_S^\top, \boldsymbol{\theta}_T^\top]^\top, \quad
\lambda_\ell\,\boldsymbol{\Theta} = 
\begin{bmatrix}
\mathbf{M}_{ST}\boldsymbol{\pi}^\top & \mathbf{M}_{SS}\boldsymbol{\pi}\\
\mathbf{M}_{TT}\boldsymbol{\pi}^\top & \mathbf{M}_{TS}\boldsymbol{\pi}
\end{bmatrix}\boldsymbol{\Theta}.
\label{eq:theta_fP_System_app}
\end{equation}

\subsection{Countable Reachability of GMM-OT coupling}
\label{app:reachability}
Consider the source and target Gaussian mixtures with plan $\pi=(\pi_{k_1,k_2})\in\Pi(\boldsymbol{\alpha}_s,\boldsymbol{\alpha}_t)$, $\sum_{k_2}\pi_{k_1,k_2}=\alpha_{sk_1}$, $\sum_{k_1}\pi_{k_1,k_2}=\alpha_{tk_2}$. For non-degenerate covariances, the optimal transport from $P_{Sk_1}$ to $P_{Tk_2}$ under the quadratic cost is the affine Monge map $T_{k_1 k_2}(x)$, and the GMM-OT coupling is defined as:
\[\gamma^*_{\small\text{GMM}}(x,y)=\sum_{k_1,k_2}\pi_{k_1,k_2}\,P_{Sk_1}(x)\,\delta_{T_{k_1 k_2}(x)}(y).\] 
Its corresponding source marginal is $\int_{\mathbb{R}^d}\gamma^*_{\mathrm{GMM}}(x,y)\,dy=\sum_{k_1}\alpha_{sk_1}P_{Sk_1}(x)=P_S(x)$. So for $P_S(x)>0$, the source-to-target conditional transition law is:
\begin{equation*}
P_{S\to T}^{\text{\tiny GMMOT}}(x,dy)=\sum_{k_1,k_2}w_{k_1 k_2}(x)\,\delta_{T_{k_1 k_2}(x)}(dy),
\end{equation*}
where,
\begin{equation*}
w_{k_1 k_2}(x)=\frac{\pi_{k_1,k_2}P_{Sk_1}(x)}{P_S(x)},\qquad \text{s.t.} \qquad \sum_{k_1,k_2}w_{k_1 k_2}(x)=1,
\end{equation*}
Despite the continuity of the mixture densities, $P_{S\to T}^{\text{\tiny GMMOT}}(x,\cdot)$ is therefore supported on finitely many points. Let $E=\{(k_1,k_2):\pi_{k_1,k_2}>0\}$ be the set of active pairs, which is finite since the mixtures have finitely many components. For any fixed $x$,
\begin{equation*}
\operatorname{supp}P_{S\to T}^{\text{\tiny GMMOT}}(x,\cdot)\subseteq\{T_{k_1 k_2}(x):(k_1,k_2)\in E\},\qquad
|\operatorname{supp}P_{S\to T}^{\text{\tiny GMMOT}}(x,\cdot)|\le|E|<\infty.
\end{equation*}
The reverse target-to-source transition is identical: the non-degenerate covariances make each $T_{k_1 k_2}$ invertible, so that reverse step is supported on the inverse images $T_{k_1 k_2}^{-1}(y)$. Thus in both directions every state has at most $M=|E|$ successors. For an initial state $x_0$, we use $R_n(x_0)$ to denote the states reachable in exactly $n$ alternating transitions and $R(x_0)=\bigcup_{n\ge0}R_n(x_0)$. Since $R_0(x_0)=\{x_0\}$ and every state has at most $M$ successors, an induction gives:
\begin{equation*}
|R_n(x_0)|\le M^n<\infty.
\end{equation*}
Hence $R(x_0)$ is a countable union of finite sets, therefore at most countable, and since every countable subset of $\mathbb{R}^d$ has zero Lebesgue measure, $\mathrm{Leb}(R(x_0))=0$.

\medskip
\subsection{Proof of Approximate Domain-Invariance}\label{app:theoretical_guarantees_proofs}
In this section, we provide the proofs and derivations to the theoretical guarantees presented in Proposition~\ref{thm:main_invariance}.
\subsubsection{Notation and semantic reference structure}
\label{subsec:notation-semantic-structure}
The theoretical analysis relies on a semantic organization of the mixture components into classes. Let $Z_S\in\{1,\ldots,K_S\}$ and $Z_T\in\{1,\ldots,K_T\}$ be the latent component variables of the source and target GMMs. For each class $C_{c_l}$ with $c_l\in\{1,\ldots,N_c\}$, the index sets $I_{S,c_l}$ and $I_{T,c_l}$ collect its source and target components. $\boldsymbol{\mathbf{(A_1)}}$ We assume every component belongs to exactly one class, so the indices split as $\{1,\ldots,K_S\}=\bigsqcup_{c_l} I_{S,c_l}$ and $\{1,\ldots,K_T\}=\bigsqcup_{c_l} I_{T,c_l}$. The class masses are $\alpha_{S,c_l}=\sum_{i\in I_{S,c_l}}\alpha_{Si}$ and $\alpha_{T,c_l}=\sum_{j\in I_{T,c_l}}\alpha_{Tj}$, with $\sum_{c_l=1}^{N_c}\alpha_{S,c_l}=\sum_{c_l=1}^{N_c}\alpha_{T,c_l}=1$. What matters here is that this assignment is fixed independently of the transport plan $\boldsymbol{\pi}$, so a component's class is one thing and where its mass ends up is another. Moreover, conditioning the latent variable on the components associated with class $c_l$ gives the class conditional densities $P_{S,c_l}(x)=\tfrac{1}{\alpha_{S,c_l}}\sum_{i\in I_{S,c_l}}\alpha_{Si} P_{Si}(x)$ and $P_{T,c_l}(x)=\tfrac{1}{\alpha_{T,c_l}}\sum_{j\in I_{T,c_l}}\alpha_{Tj} P_{T
j}(x)$.
\paragraph{High density regions.} For a density threshold $\xi >0$ we look at the high density regions $\Omega_{S,c_l,\xi}=\{x:P_{S,c_l}(x)\geq\xi\}$ and $\Omega_{T,c_l,\xi}=\{x:P_{T,c_l}(x)\geq\xi\}$. $\mathbf{(A_2)}$ Throughout we assume there is some $\xi>0$ for which these cores are pairwise disjoint inside each domain, meaning $\Omega_{S,c_l,\xi}\cap\Omega_{S,c_l',\xi}=\varnothing$ and $\Omega_{T,c_l,\xi}\cap\Omega_{T,c_l',\xi}=\varnothing$ whenever $c_l\neq c_l'$. Notice this only separates the high density regions and does not ask the full supports to be disjoint. Finally we define the cross domain core $\Omega_{c_l,\xi}=\Omega_{S,c_l,\xi}\cup\Omega_{T,c_l,\xi}$, which is just an ordinary union since the source and target pieces live in the same space and may well overlap.
\paragraph{Transport Mismatches.} Taking into consideration that the OT problem is blind to class labels, it does not prevent assigning mass between source and target components that are associated to different classes, this introduces an error at the level of the OT plan that is denoted by:
\begin{equation}
\ell_{S,c_l, i}^{\mathrm{OT}}=\frac{1}{\alpha_{Si}}\sum_{j\notin I_{T,c_l}}\pi_{ij},\qquad
\ell_{T,c_l, j}^{\mathrm{OT}}=\frac{1}{\alpha_{Tj}}\sum_{i\notin I_{S,c_l}}\pi_{ij},
\label{eq:local-leakage}
\end{equation}
for a source component $i\in I_{S,c_l}$ and a target component $j\in I_{T,c_l}$. These quantities measure the fraction of the marginal mass of an individual mixture component that is transported toward components associated with other classes. By the marginal constraints, this fixes the complementary mass that stays inside the class block,
\begin{equation}
\sum_{j\in I_{T,c_l}}\pi_{ij}=\alpha_{Si}\bigl(1-\ell_{S,c_l, i}^{\mathrm{OT}}\bigr),\qquad
\sum_{i\in I_{S,c_l}}\pi_{ij}=\alpha_{Tj}\bigl(1-\ell_{T,c_l, j}^{\mathrm{OT}}\bigr).
\label{eq:leakage-marginals}
\end{equation}
Note that if we aggregate over a class we obtain the class level mismatches: \[
L_{S,c_l}=\sum_{i\in I_{S,c_l}}\alpha_{Si}\ell_{S,c_l, i}^{\mathrm{OT}}=\sum_{\substack{i\in I_{S,c_l}\\ j\notin I_{T,c_l}}}\pi_{ij}, \qquad L_{T,c_l}=\sum_{j\in I_{T,c_l}}\alpha_{Tj}\ell_{T,c_l, j}^{\mathrm{OT}}=\sum_{\substack{i\notin I_{S,c_l}\\ j\in I_{T,c_l}}}\pi_{ij}.
\]

\subsubsection{Bound on the intra-class transport plan and continuous coupling}
\label{subsec:intra-class-bound}
In this part we derive a pointwise lower bound on every intra-class transport coefficient. We start from the Entropic OT problem discussed in Section \ref{sec:background_ot}.
\begin{tcolorbox}[colback=gray!8, colframe=gray!35, boxrule=0.2pt, arc=1pt, left=1pt, right=1pt, top=1pt, bottom=1pt]
\begin{proposition}[Lower bound on intra-class transport coefficients]
\label{prop:intra_class_transport_lower_bound}
Under the Sinkhorn factorization $\pi_{ij}=u_iK_{ij}v_j$ with $K_{ij}=e^{-C_{ij}/\varepsilon}$ and $u_i,v_j>0$, worst-case leakages $\ell_{S,c_l}^{\mathrm{OT}},\ell_{T,c_l}^{\mathrm{OT}}<1$, and balanced masses $m_{c_l}=\alpha_{S,c_l}=\alpha_{T,c_l}$, every $i\in I_{S,c_l}$, $j\in I_{T,c_l}$ satisfies:
\begin{equation}
\pi_{ij}\geq\frac{\kappa_{c_l}^2}{m_{c_l}}\bigl(1-\ell_{S,c_l}^{\mathrm{OT}}\bigr)\bigl(1-\ell_{T,c_l}^{\mathrm{OT}}\bigr)\alpha_{Si}\alpha_{Tj},
\label{eq:explicit_intra_class_pi_lower_bound}
\end{equation}
equivalently
\begin{equation}
\frac{\pi_{ij}}{\alpha_{Si}\alpha_{Tj}}\geq\frac{e^{-2\Delta C_{c_l}/\varepsilon}}{m_{c_l}}\bigl(1-\ell_{S,c_l}^{\mathrm{OT}}\bigr)\bigl(1-\ell_{T,c_l}^{\mathrm{OT}}\bigr).
\label{eq:relative_intra_class_pi_lower_bound}
\end{equation}
\end{proposition}
\end{tcolorbox}
\begin{proof}
 We fix a semantic class $c_l$ and denote, for simplicity, $I=I_{S,c_l},\quad J=I_{T,c_l}$. Throughout we use the Sinkhorn factorization $\pi_{ij}=u_iK_{ij}v_j$ with positive scaling factors $u_i,v_j$ and $K_{ij}=\exp(-C_{ij}/\varepsilon)$. We denote the transport mass retained by row $i\in I$ and by column $j\in J$ inside the class block according to \ref{eq:leakage-marginals}. Let $K^{\max}_{c_l}=\max_{\substack{i\in I\\j\in J}}K_{ij}$ and $ K^{\min}_{c_l}=\min_{\substack{i\in I\\j\in J}}K_{ij}$. Since $K_{ij}=\exp(-C_{ij}/\varepsilon)$, we have:
\begin{equation}
\frac{K^{\min}_{c_l}}{K^{\max}_{c_l}}
=\exp\!\Bigl(-\tfrac{C_{c_l}^{\max}-C_{c_l}^{\min}}{\varepsilon}\Bigr)
=\exp\!\Bigl(-\tfrac{\Delta C_{c_l}}{\varepsilon}\Bigr)
=\kappa_{c_l} \implies K^{\min}_{c_l}=\kappa_{c_l}\,K^{\max}_{c_l}.
\label{app:eq:gibbs_min_max_relation}
\end{equation}
\paragraph{Lower bounds on the Sinkhorn scaling factors.} We define $U_{c_l}=\sum_{i\in I}u_i$ and $V_{c_l}=\sum_{j\in J}v_j$. Using factorization $\sum_{j\in I_{T,c_l}}\pi_{ij}=\sum_{j\in J}u_iK_{ij}v_j=u_i\sum_{j\in J}K_{ij}v_j$, and since $K_{ij}\le K^{\max}_{c_l}$ within the class block, we have the following:
\begin{equation}
\sum_{j\in I_{T,c_l}}\pi_{ij}\le u_iK^{\max}_{c_l}V_{c_l} \implies u_i\ge\frac{\sum_{j\in I_{T,c_l}}\pi_{ij}}{K^{\max}_{c_l}V_{c_l}}.
\label{app:eq:source_scaling_lower_bound}
\end{equation}
Similarly,
\begin{align}
\sum_{i\in I_{S,c_l}}\pi_{ij}&=\sum_{i\in I}u_iK_{ij}v_j=v_j\sum_{i\in I}u_iK_{ij}\le v_jK^{\max}_{c_l}U_{c_l}, \nonumber\\
& \implies v_j\ge\frac{\sum_{i\in I_{S,c_l}}\pi_{ij}}{K^{\max}_{c_l}U_{c_l}}.
\label{app:eq:target_scaling_lower_bound}
\end{align}
\paragraph{Control of the total scaling inside the class block.} Let $\Gamma_{c_l}=\sum_{i\in I}\sum_{j\in J}\pi_{ij}$ denote the total transport mass retained inside the $c_l$-th class block. Using again the factorization, one obtains:
\[
\Gamma_{c_l}=\sum_{i\in I}\sum_{j\in J}u_iK_{ij}v_j
\ge K^{\min}_{c_l}\Bigl(\sum_{i\in I}u_i\Bigr)\Bigl(\sum_{j\in J}v_j\Bigr)
=K^{\min}_{c_l}U_{c_l}V_{c_l}.
\]
Using \eqref{app:eq:gibbs_min_max_relation}, $\Gamma_{c_l}\ge\kappa_{c_l}K^{\max}_{c_l}U_{c_l}V_{c_l}$, therefore:
\begin{equation}
K^{\max}_{c_l}U_{c_l}V_{c_l}\le\frac{\Gamma_{c_l}}{\kappa_{c_l}}.
\label{app:eq:scaling_product_upper_bound}
\end{equation}
Under the balanced class-mass assumption $\sum_{i\in I}\alpha_{Si}=\sum_{j\in J}\alpha_{Tj}=m_{c_l}$, and since $\Gamma_{c_l}$ is only the part of the class marginal mass that remains inside the block,
\begin{equation}
\Gamma_{c_l}=\sum_{i\in I}\sum_{j\in J}\pi_{ij}\le m_{c_l}.
\label{app:eq:class_block_mass_upper_bound}
\end{equation}
\paragraph{Pointwise lower bound on the OT coupling.} For any $i\in I$ and $j\in J$, one has:
\begin{align}
\pi_{ij}=u_iK_{ij}v_j
&\overset{\text{(a)}}{\ge}\frac{\bigl(\sum_{j\in I_{T,c_l}}\pi_{ij}\bigr)\bigl(\sum_{i\in I_{S,c_l}}\pi_{ij}\bigr)K_{ij}}{(K^{\max}_{c_l})^2U_{c_l}V_{c_l}}\nonumber
\\&\overset{\text{(b)}}{\ge}\frac{\kappa_{c_l}\bigl(\sum_{j\in I_{T,c_l}}\pi_{ij}\bigr)\bigl(\sum_{i\in I_{S,c_l}}\pi_{ij}\bigr)}{K^{\max}_{c_l}U_{c_l}V_{c_l}}\nonumber
\\&\overset{\text{(c)}}{\ge}\frac{\kappa_{c_l}^2\bigl(\sum_{j\in I_{T,c_l}}\pi_{ij}\bigr)\bigl(\sum_{i\in I_{S,c_l}}\pi_{ij}\bigr)}{\Gamma_{c_l}} \nonumber
\\&\overset{\text{(d)}}{\ge}\frac{\kappa_{c_l}^2}{m_{c_l}}\bigl(\sum_{j\in I_{T,c_l}}\pi_{ij}\bigr)\bigl(\sum_{i\in I_{S,c_l}}\pi_{ij}\bigr)
\nonumber\\
&\overset{\text{(e)}}{=}\frac{\kappa_{c_l}^2}{m_{c_l}}\,\alpha_{Si}\alpha_{Tj}\bigl(1-\ell_{S,c_l, i}^{\mathrm{OT}}\bigr)\bigl(1-\ell_{S,c_l, i}^{\mathrm{OT}}\bigr)\nonumber
\\&\overset{\text{(f)}}{\ge}\frac{\kappa_{c_l}^2}{m_{c_l}}\bigl(1-\ell_{S,c_l}^{\mathrm{OT}}\bigr)\bigl(1-\ell_{T,c_l}^{\mathrm{OT}}\bigr)\alpha_{Si}\alpha_{Tj},
\label{app:eq:explicit_intra_class_pi_lower_bound}
\end{align}
where (a) inserts the scaling lower bounds \eqref{app:eq:source_scaling_lower_bound} and \eqref{app:eq:target_scaling_lower_bound} into the factorization, (b) uses $K_{ij}\ge K^{\min}_{c_l}=\kappa_{c_l}K^{\max}_{c_l}$ from \eqref{app:eq:gibbs_min_max_relation}, (c) applies the total-scaling bound \eqref{app:eq:scaling_product_upper_bound}, (d) uses $\Gamma_{c_l}\le m_{c_l}$ from \eqref{app:eq:class_block_mass_upper_bound}, (e) substitutes the retained masses \eqref{eq:leakage-marginals}, and (f) bounds each local leakage by its worst-case value, $\ell_{S,c_l, i}^{\mathrm{OT}}\le\ell_{S,c_l}^{\mathrm{OT}}$ and $\ell_{T,c_l, j}^{\mathrm{OT}}\le\ell_{T,c_l}^{\mathrm{OT}}$.
\end{proof}
\subsubsection{Continuous coupling and affinity kernel}\label{bound on affinity kernel}
Here, we use the result on Proposition~\ref{prop:intra_class_transport_lower_bound} to move from the discrete component coupling to the continuous dynamics. For a class $c_l$, we define the intra-class contribution as:
\begin{align}
\pi_{c_l}(x,y)=\sum_{i\in I_{S,c_l}}\sum_{j\in I_{T,c_l}}\pi_{ij}P_{si}(x)P_{tj}(y).
\label{eq:class_continuous_coupling}
\end{align}
Applying the bound of $\pi_{ij}$ in \ref{app:eq:explicit_intra_class_pi_lower_bound}, we obtain:
\begin{align}
\pi_{c_l}(x,y)
&\geq
\frac{\kappa_{c_l}^2}{m_{c_l}}
\left(1-\ell_{S,c_l}^{\mathrm{OT}}\right)
\left(1-\ell_{T,c_l}^{\mathrm{OT}}\right)
\times
\left(
\sum_{i\in I_{S,c_l}}
\alpha_{Si}P_{Si}(x)
\right)
\left(
\sum_{j\in I_{T,c_l}}
\alpha_{Tj} P_{Tj}(y)
\right)
\end{align}
Since $\pi(x,y) \ge \pi_{c_l}(x,y)$ (because all terms of the full coupling are nonnegative), and using that $P_{S,c_l}(x)=\tfrac{1}{\alpha_{S,c_l}}\sum_{i\in I_{S,c_l}}\alpha_{Si} P_{Si}(x)$ and $P_{T,c_l}(x)=\tfrac{1}{\alpha_{T,c_l}}\sum_{j\in I_{T,c_l}}\alpha_{Tj} P_{T
j}(x)$ together with the definition of high density regions for which $p_{S,c_l}(x)\geq\xi$ and$p_{T,c_l}(y)\geq\xi,$ gives:
\begin{equation}
\pi(x,y)
\geq
m_{c_l}\xi^2
e^{-2\Delta C_{c_l}/\varepsilon}
\left(1-\ell_{S,c_l}^{\mathrm{OT}}\right)
\left(1-\ell_{T,c_l}^{\mathrm{OT}}\right).
\label{eq:continuous_core_connectivity_explicit}
\end{equation}
Propagating this bound to the symmetric affinity kernel such that $\mathcal{A}(x,y)
=\frac12\left[\pi(x,y)+\pi(y,x)\right]$, given that both terms are nonnegative, one obtains:
\begin{equation}
\resizebox{\linewidth}{!}{$\displaystyle
\mathcal{A}(x,y)
\geq
\beta_{c_l,\xi} \quad \text{s.t.,} \quad \beta_{c_l,\xi}
=\frac12
m_{c_l}\xi^2
e^{-2\Delta C_{c_l}/\varepsilon}
\left(1-\ell_{S,{c_l}}^{\mathrm{OT}}\right)
\left(1-\ell_{T,c_l}^{\mathrm{OT}}\right),
\quad
\forall(x,y)\in
\Omega_{S,c_l,\xi}
\times
\Omega_{T,c_l,\xi}
$}
\label{eq:affinity bound}
\end{equation}
\subsubsection{Slow-mode stability over cross-domain class cores}
We now establish the main connection between the metastable interpretation of the cross-domain dynamics and domain invariance of its slow spectral modes. Let $f_\ell$ be an eigenfunction of $\mathcal{K}$ associated with an eigenvalue $\lambda_\ell$  \textit{s.t.} $\mathcal{K}f_\ell=\lambda_\ell f_\ell$.
The Dirichlet form associated with $\mathcal{K}$ is
$\mathcal E(f,h)=\langle (I-\mathcal{K})f,h\rangle_d$. Therefore:
\begin{equation}
 \mathcal E(f_\ell,f_\ell)
 =(1-\lambda_\ell)\|f_\ell\|_d^2.
 \label{eq:eigenergy}
\end{equation}
On the other hand, and using the symmetry of $\mathcal{A}(x,y)$, we have:
\begin{align}
\mathcal{E}(f_\ell, f_\ell)
&= \int_{\mathbb{R}^d} \big( f_\ell(x) - \mathcal{K}[f_\ell](x) \big)\, f_\ell(x)\, d(x)\, dx \nonumber\\
&= \int_{\mathbb{R}^d} \Big( f_\ell(x) - \int_{\mathbb{R}^d} p(x,y) f_\ell(y)\, dy \Big)\, f_\ell(x)\, d(x)\, dx \nonumber\\
&= \int_{\mathbb{R}^d} \Big( {\int_{\mathbb{R}^d} f_\ell(x) p(x,y)\, dy} - \int_{\mathbb{R}^d} p(x,y) f_\ell(y)\, dy \Big)\, f_\ell(x)\, d(x)\, dx \nonumber\\
&= \int_{\mathbb{R}^d} \Big( \int_{\mathbb{R}^d} p(x,y) \big( f_\ell(x) - f_\ell(y) \big)\, dy \Big)\, f_\ell(x)\, d(x)\, dx \nonumber\\
&= \iint_{\mathbb{R}^d \times \mathbb{R}^d} d(x)\, p(x,y) \big( f_\ell(x)^2 - f_\ell(x) f_\ell(y) \big)\, dy\, dx \nonumber\\
&= \iint_{\mathbb{R}^d \times \mathbb{R}^d} \mathcal{A}(x,y) \big( f_\ell(x)^2 - f_\ell(x) f_\ell(y) \big)\, dy\, dx \nonumber\\
&= \frac{1}{2} \iint_{\mathbb{R}^d \times \mathbb{R}^d} \mathcal{A}(x,y) \big( f_\ell^2(x) - f_\ell(x) f_\ell(y) \big)
+ \mathcal{A}(y,x) \big( f_\ell^2(y) - f_\ell(y) f_\ell(x) \big)\, dy\, dx \nonumber \\
&= \frac{1}{2} \iint_{\mathbb{R}^d \times \mathbb{R}^d} \mathcal{A}(x,y) \big( f_\ell^2(x) - 2 f_\ell(x) f_\ell(y) + f_\ell^2(y) \big)\, dy\, dx \nonumber\\
&= \frac{1}{2} \iint_{\mathbb{R}^d \times \mathbb{R}^d} \mathcal{A}(x,y) \big( f_\ell(x) - f_\ell(y) \big)^2\, dy\, dx
\label{eq:dirichlet_energy}
\end{align}
Combining this identity with \eqref{eq:eigenergy} yields:
\begin{equation}
 \iint \mathcal{A}(x,y)\bigl(f_\ell(x)-f_\ell(y)\bigr)^2\,dx\,dy
 =2(1-\lambda_\ell)\|f_\ell\|_d^2.
 \label{eq:global-energy}
\end{equation}
Thus, for $\lambda_\ell$ close to one, $f_\ell$ has small quadratic variation along transitions carrying a large affinity. The key observation here is that the Dirichlet energy of an eigenfunction associated with an eigenvalue close to one is small. We now combine this with the uniform lower bound on the affinity kernel established in Section~\ref{bound on affinity kernel}. For a class $c_l$, let $ S_{c_l}=\Omega_{S,c_l,\xi}$ and $T_{c_l}=\Omega_{T,c_l,\xi}$ and denote their Lebesgue volumes by $V_{S,c_l}=|S_{c_l}|$ and $V_{T,c_l}=|T_{c_l}|$ respectively. Restricting the Dirichlet integral in
Eq.~\eqref{eq:dirichlet_energy} to $S_{c_l}\times T_{c_l}$ yields:\\
\begin{equation}
    \mathcal{E}(f_\lambda,f_\lambda)
    \geq
    \frac{1}{2}
    \int_{S_{c_l}}
    \int_{T_{c_l}}
    \mathcal{A}(x,y)
    \left(
        f_\lambda(x)-f_\lambda(y)
    \right)^2
    dy\,dx.
\end{equation}
Using \eqref{eq:affinity bound}, one gets:
\begin{equation}
    \mathcal{E}(f_\lambda,f_\lambda)
    \geq
    \frac{\beta_{c_l,\xi}}{2}
    \int_{S_{c_l}}
    \int_{T_{c_l}}
    \left(
        f_\lambda(x)-f_\lambda(y)
    \right)^2
    dy\,dx.
    \label{app:eq:energy_controls_cross_variation}
\end{equation}
\begin{equation}
    \int_{S_{c_l}}
    \int_{T_{c_l}}
    \left(
        f_\lambda(x)-f_\lambda(y)
    \right)^2
    dy\,dx
    \leq
    \frac{
        2(1-\lambda)
    }{
        \beta_{c_l,\xi}
    }
    \|f_\lambda\|_{L^2(\nu)}^2.
    \label{app:eq:cross_variation_bound}
\end{equation}
\subsubsection{Variance bound over the union of cores}
We now relate the cross-domain variation in Eq.~\eqref{app:eq:cross_variation_bound} directly to the variability of $f_\lambda$ over the union$U_{c_l}=S_{c_l}\cup T_{c_l}$. Importantly, no disjointness assumption is made between $S_{cl}$ and $T_{cl}$. We define $V_{cl}^{\min}=\min\left\{
V_{S,c_l}, V_{T,c_l}\right\}$. Let the uniform average of $f_\lambda$ over the target core be defined as:
\begin{equation}
  \tilde{f}_{T,c_l}=\frac{1}{V_{T,c_l}}\int_{T_{c_l}}f_\lambda(y)\,dy.  
\end{equation}
For any fixed $x\in S_{c_l}$, expanding around $\tilde{f}_{T,c_l}$ gives:
\begin{align}
    &
    \int_{T_{c_l}}
    \left(
        f_\lambda(x)-f_\lambda(y)
    \right)^2
    dy
=
    \int_{T_{c_l}}
    \left[
        \left(
            f_\lambda(x)-\tilde{f}_{T,c_l}
        \right)
        +
        \left(
            \tilde{f}_{T,c_l}-f_\lambda(y)
        \right)
    \right]^2
    dy. \\
    &\resizebox{0.9\linewidth}{!}{$\displaystyle
    =\big(
    V_{T,c_l}
    \left(
        f_\lambda(x)-\tilde{f}_{T,c_l}
    \right)^2 +
    \int_{T_{c_l}}
    \left(
        f_\lambda(y)-\tilde{f}_{T,c_l}
    \right)^2
    dy
+
    2
    \left(
        f_\lambda(x)-\tilde{f}_{T,c_l}
    \right)
    \int_{T_{c_l}}
    \left(
        \tilde{f}_{T,c_l}-f_\lambda(y)
    \right)
    dy\big).$}
\end{align}
By definition of $\tilde{f}_{T,c_l}$,
\begin{equation}
    \int_{T_{c_l}}
    \left(
        \tilde{f}_{T,c_l}-f_\lambda(y)
    \right)
    dy
    =
    0.
\end{equation}
Therefore,
\begin{align}
    &
    \int_{T_{c_l}}
    \left(
        f_\lambda(x)-f_\lambda(y)
    \right)^2
    dy
=
    V_{T,c_l}
    \left(
        f_\lambda(x)-\tilde{f}_{T,c_l}
    \right)^2
    +
    \int_{T_{c_l}}
    \left(
        f_\lambda(y)-\tilde{f}_{T,c_l}
    \right)^2
    dy.
    \label{app:eq:pointwise_cross_variation_decomposition}
\end{align}
Integrating Eq.~\eqref{app:eq:pointwise_cross_variation_decomposition}
over $x\in S_{c_l}$ gives the exact identity
\begin{align}
    \int_{S_{c_l}}
    \int_{T_{c_l}}
    \left(
        f_\lambda(x)-f_\lambda(y)
    \right)^2
    dy\,dx
    =
    &V_{T,c_l}
    \int_{S_{c_l}}
    \left(
        f_\lambda(x)-\tilde{f}_{T,c_l}
    \right)^2
    dx \nonumber \\
&+
    V_{S,c_l}
    \int_{T_{c_l}}
    \left(
        f_\lambda(y)-\tilde{f}_{T,c_l}
    \right)^2
    dy.
    \label{app:eq:cross_variation_decomposition}
\end{align}
Using the definition of $V_{cl}^{\min}$,
Eq.~\eqref{app:eq:cross_variation_decomposition} implies:
\begin{align}
    & \resizebox{0.92\linewidth}{!}{$\displaystyle
    \int_{S_{c_l}}
    \int_{T_{c_l}}
    \left(
        f_\lambda(x)-f_\lambda(y)
    \right)^2
    dy\,dx
 \geq
    V_{cl}^{\min}
    \left[
        \int_{S_{c_l}}
        \left(
            f_\lambda(x)-\tilde{f}_{T,c_l}
        \right)^2
        dx
    \right.
    \left.
        +
        \int_{T_{c_l}}
        \left(
            f_\lambda(y)-\tilde{f}_{T,c_l}
        \right)^2
        dy
    \right]. $}
    \label{app:eq:cross_variation_min_volume}
\end{align}
Because the integrand is nonnegative and
$U_{c_l}=S_{c_l}\cup T_{c_l}$, the sum of the two integrals on the right-hand side
counts points in $S_{c_l}\cap T_{c_l}$ twice and therefore satisfies:
\begin{align}
    &
    \int_{S_{c_l}}
    \left(
        f_\lambda-\tilde{f}_{T,c_l}
    \right)^2
    +
    \int_{T_{c_l}}
    \left(
        f_\lambda-\tilde{f}_{T,c_l}
    \right)^2
\geq
    \int_{U_{c_l}}
    \left(
        f_\lambda(z)-\tilde{f}_{T,c_l}
    \right)^2
    dz.
    \label{app:eq:union_froN_Sum}
\end{align}
We now define the uniform mean of $f_\lambda$ over the actual union
$U_{c_l}$ as:
\begin{equation}
    \tilde{f}_{U,c_l}
    =
    \frac{1}{|U_{c_l}|}
    \int_{U_{c_l}}
    f_\lambda(z)\,dz.
    \label{app:eq:union_average}
\end{equation}
The mean $\tilde{f}_{U,c_l}$ minimizes the quadratic deviation over
$U_{c_l}$. Hence, for any constant $c$,
\begin{equation}
    \int_{U_{c_l}}
    \left(
        f_\lambda(z)-\tilde{f}_{U,c_l}
    \right)^2
    dz
    \leq
    \int_{U_{c_l}}
    \left(
        f_\lambda(z)-c
    \right)^2
    dz.
\end{equation}
Choosing $c=\tilde{f}_{T,c_l}$ gives
\begin{equation}
    \int_{U_{c_l}}
    \left(
        f_\lambda(z)-\tilde{f}_{T,c_l}
    \right)^2
    dz
    \geq
    \int_{U_{c_l}}
    \left(
        f_\lambda(z)-\tilde{f}_{U,c_l}
    \right)^2
    dz.
    \label{app:eq:mean_minimizes_variance}
\end{equation}
Define the integrated variance over the union by:
\begin{equation}
    \mathcal{V}_{U_{c_l}}(f_\lambda)
    =
    \int_{U_{c_l}}
    \left(
        f_\lambda(z)-\tilde{f}_{U,c_l}
    \right)^2
    dz.
    \label{app:eq:integrated_union_variance}
\end{equation}
Combining
Eqs.~\eqref{app:eq:cross_variation_min_volume},
\eqref{app:eq:union_froN_Sum},
and~\eqref{app:eq:mean_minimizes_variance}, we obtain:
\begin{equation}
    \int_{S_{c_l}}
    \int_{T_{c_l}}
    \left(
        f_\lambda(x)-f_\lambda(y)
    \right)^2
    dy\,dx
    \geq
    V_{cl}^{\min}
    \mathcal{V}_{U_{c_l}}(f_\lambda).
    \label{app:eq:cross_variation_controls_union_variance}
\end{equation}
Equivalently,
\begin{equation}
    \mathcal{V}_{U_{c_l}}(f_\lambda)
    \leq
    \frac{1}{V_{cl}^{\min}}
    \int_{S_{c_l}}
    \int_{T_{c_l}}
    \left(
        f_\lambda(x)-f_\lambda(y)
    \right)^2
    dy\,dx.
    \label{app:eq:union_variance_from_cross_variation}
\end{equation}
Combining
Eq.~\eqref{app:eq:union_variance_from_cross_variation}
with Eq.~\eqref{app:eq:cross_variation_bound}
yields the slow-mode stability bound.

\subsection{On the Estimation of The Overlap matrices}\label{overlap_matrices_mc}
\begin{tcolorbox}[colback=gray!8, colframe=gray!35, boxrule=0.4pt, arc=2pt, left=2pt, right=2pt, top=2pt, bottom=2pt]
\begin{proposition}[Monte Carlo estimation of overlap matrices]\label{prop:mc_overlaps}
Each entry of the overlap matrices admits an unbiased Monte Carlo estimator. For $x_s^{(n)} \sim P_{Si}$ and $x_t^{(n)} \sim P_{Tj}$, 
drawn independently for $n = 1, \dots, N$,
\begin{equation}
\widehat{M}_{SS}[i,k_1] = \frac{1}{N}\sum_{n=1}^N 
\frac{P_{Sk_1}(x_s^{(n)})}{P_S(x_s^{(n)})+P_T(x_s^{(n)})}, \quad
\widehat{M}_{ST}[i,k_2] = \frac{1}{N}\sum_{n=1}^N 
\frac{P_{Tk_2}(x_s^{(n)})}{P_S(x_s^{(n)})+P_T(x_s^{(n)})},
\end{equation}
\begin{equation}
\widehat{M}_{TS}[j,k_1] = \frac{1}{N}\sum_{n=1}^N 
\frac{P_{Sk_1}(x_t^{(n)})}{P_S(x_t^{(n)})+P_T(x_t^{(n)})}, \quad
\widehat{M}_{TT}[j,k_2] = \frac{1}{N}\sum_{n=1}^N 
\frac{P_{Tk_2}(x_t^{(n)})}{P_S(x_t^{(n)})+P_T(x_t^{(n)})},
\end{equation}
all converging almost surely as $N \to \infty$ to their population counterparts $M_{SS}[i,k_1] = \mathbb{E}_{x\sim P_{Si}}[g_{Sk_1}(x)]$, $M_{ST}[i,k_2] = \mathbb{E}_{x\sim P_{Si}}[g_{Tk_2}(x)]$, $M_{TS}[j,k_1] = \mathbb{E}_{x\sim P_{Tj}}[g_{Sk_1}(x)]$, and $M_{TT}[j,k_2] = \mathbb{E}_{x\sim P_{Tj}}[g_{Tk_2}(x)]$.
\end{proposition}
\end{tcolorbox}
In the isotropic lifting special case, since each Gaussian component reads $P_{Si} = \mathcal{N}(\mathbf{x}_{Si}, \boldsymbol{\sigma}^2I)$ and $P_{Tj} = \mathcal{N}(\mathbf{x}_{Tj}, \boldsymbol{\sigma}^2I)$, sampling from them reduces to the reparameterization $x_s^{(n)} = \mathbf{x}_{Si} + \sigma\,\epsilon^{(n)}$ and $x_t^{(n)} = \mathbf{x}_{Tj} + \sigma\,\epsilon^{(n)}$ with $\epsilon^{(n)} \sim \mathcal{N}(0, I_d)$. The normalized weights $g_{Sk_1}(x_s^{(n)})$ and $g_{Tk_2}(x_s^{(n)})$ appearing in the estimators are then evaluated in log-space for numerical stability. All four matrices are estimated once and kept fixed throughout, since they depend only on the lifted distributions $P_S$ and $P_T$ and not on the learnable parameters $\boldsymbol{\theta}_S$ and $\boldsymbol{\theta}_T$.

\subsection{NeuralSPOT Baseline}\label{App:neuralspot}

Let $\boldsymbol{X_s} = \{\mathbf{x_s}^i\}_{i=1}^{n_s}$ and $\boldsymbol{X_t} = \{\mathbf{x_t}^j\}_{j=1}^{n_t}$
be two empirical distributions supported on source and target probability measures $\mu_s$ and $\mu_t$, respectively, with $n_s \neq n_t$. Given an entropic optimal transport plan $\boldsymbol{\pi_{\varepsilon}}$ between these empirical measures, we follow the formulation of \citep{sadsaoud2026beyond}, where source and target samples are 
represented as nodes of a bipartite graph whose adjacency matrix is:
\begin{equation}\label{adjacency_discrete}
\mathcal{A} = \begin{pmatrix} 0 & \boldsymbol{\pi_{\varepsilon}} \\ \boldsymbol{\pi_{\varepsilon}}^{\top} & 0 \end{pmatrix} 
\in \mathbb{R}^{(n_s + n_t) \times (n_s + n_t)},
\end{equation}
and the corresponding degree matrix is:
\begin{equation}\label{degree_discrete}
\boldsymbol{D} = \begin{pmatrix} \boldsymbol{D_s} & 0 \\ 0 & \boldsymbol{D_t} \end{pmatrix},
\quad
(D_s)_{ii} = \sum_{j=1}^{n_t} \pi_{{\varepsilon}_{ij}},
\quad
(D_t)_{jj} = \sum_{i=1}^{n_s} \pi_{{\varepsilon}_{ij}}.
\end{equation}
The corresponding random-walk Laplacian is defined as $\boldsymbol{L}_{\mathrm{rw}} = \boldsymbol{D^{-1}L} = \boldsymbol{I_{n_s + n_t}} 
- \boldsymbol{D^{-1}}\mathcal{A}$.
\begin{tcolorbox}[colback=gray!8, colframe=gray!35, boxrule=0.4pt, arc=2pt, left=2pt, right=2pt, top=2pt, bottom=2pt]
\begin{lemma}
Let $\mathcal{A}$ and $\boldsymbol{D}$ be defined as above. The matrix $\boldsymbol{P} = 
\boldsymbol{D^{-1}}\mathcal{A}$ is row-stochastic and defines the transition operator 
of a discrete-time Markov chain on the bipartite graph with adjacency 
$\mathcal{A}$:
\begin{equation}\label{prob_transition_operator}
\boldsymbol{P} = \boldsymbol{D^{-1}}\mathcal{A} = \begin{pmatrix} 0 & \boldsymbol{D_s^{-1}\pi_{\varepsilon}} \\ 
\boldsymbol{D_t^{-1}\pi_{\varepsilon}}^{\top} & 0 \end{pmatrix}, 
\quad \boldsymbol{L_{\mathrm{rw}}} = \boldsymbol{I_{n_s + n_t}} - \boldsymbol{P}.
\end{equation}
\end{lemma}
\end{tcolorbox}
The spectral embedding $\boldsymbol{\mathbf{Z}} \in \mathbb{R}^{(n_s + n_t) \times k}$ 
of source and target samples reads for each 
column $\boldsymbol{z}$ of $\boldsymbol{\mathbf{Z}}$,
\begin{equation}\label{eigen-decomp}
\boldsymbol{L_{\mathrm{rw}} z} = 0 \iff \boldsymbol{Pz} = \boldsymbol{z},
\end{equation}
so $\boldsymbol{z}$ is invariant under one transition step of $\boldsymbol{P}$, equivalently it lies in the stationary subspace of the Markov chain. Decomposing 
$\boldsymbol{\mathbf{Z}}$ into source and target blocks $\boldsymbol{\mathbf{Z}_s}$ and 
$\boldsymbol{\mathbf{Z}_t}$,
\begin{equation}
\begin{pmatrix} 0 & \boldsymbol{D_s^{-1}\pi_{\varepsilon}} \\ \boldsymbol{D_t^{-1}\pi_{\varepsilon}}^{\top} & 0 \end{pmatrix}
\begin{pmatrix} \boldsymbol{\mathbf{Z}_s} \\ \boldsymbol{\mathbf{Z}_t} \end{pmatrix} 
= \begin{pmatrix} \boldsymbol{\mathbf{Z}_s} \\ \boldsymbol{\mathbf{Z}_t} \end{pmatrix},
\end{equation}
which yields the coupled eigen-problem
\begin{equation}\label{coupled_eigen}
\begin{cases}
\boldsymbol{D_s^{-1}\pi_{\varepsilon}\,\mathbf{Z}_t} = \boldsymbol{\mathbf{Z}_s}, \\
\boldsymbol{D_t^{-1}\pi_{\varepsilon}{\top}\,\mathbf{Z}_s} = \boldsymbol{\mathbf{Z}_t}.
\end{cases}
\end{equation}
\begin{tcolorbox}[colback=gray!8, colframe=gray!35, boxrule=0.4pt, arc=2pt, left=2pt, right=2pt, top=2pt, bottom=2pt]
\begin{proposition}
Let $\Phi : \mathbb{R}^d \to \mathbb{R}^k$, $f_\Phi = f_{\Phi_l} 
\circ \dots \circ f_{\Phi_1}$, be a neural network with learnable 
parameters $\Phi = \{\Phi_1, \dots, \Phi_l\}$, where $k$ is the 
number of eigenvectors. The coupled eigen-problem~\eqref{coupled_eigen} 
can be solved by learning a shared embedding $\mathbf{Z}_s[i] = 
f_\Phi(\mathbf{x_i^s})$ and $\mathbf{Z}_t[j] = f_\Phi(\mathbf{x_j^t})$ that minimizes
\begin{equation}\label{neuralspot_spectral_loss}
\mathcal{L}_{\mathrm{spec}}(\Phi) = \textcolor{black}{\left\|D_s^{-1}\pi_\varepsilon\, F_\Phi(\mathbf{X_t}) - F_\Phi(\mathbf{X_s})\right\|_F^2
+ \left\|D_t^{-1}\pi_\varepsilon{\top}\, F_\Phi(\mathbf{X_s}) - F_\Phi(\mathbf{X_t})\right\|_F^2,}
\end{equation}

\textcolor{black}{where $F_\Phi(\mathbf{X_s}) = [f_\Phi(\mathbf{x_1^s}); \dots; f_\Phi(\mathbf{x_{n_s}^s})] \in \mathbb{R}^{n_s\times k}$ and $F_\Phi(\mathbf{X_t})$ is defined analogously, i.e.,}
\begin{equation}
\mathcal{L}_{\mathrm{spec}}(\Phi) = \textcolor{black}{\sum_{i=1}^{n_s}\Big\|\sum_{j=1}^{n_t}} (D_s^{-1}\pi_\varepsilon)_{ij}\, f_\Phi(\mathbf{x_j^t}) - f_\Phi(\mathbf{x_i^s})\textcolor{black}{\Big\|_2^2}
+ \textcolor{black}{\sum_{j=1}^{n_t}\Big\|\sum_{i=1}^{n_s}} (D_t^{-1}\pi_\varepsilon{\top})_{ji}\, f_\Phi(\mathbf{x_i^s}) - f_\Phi(\mathbf{x_j^t})\textcolor{black}{\Big\|_2^2}.
\end{equation}
\end{proposition}
\end{tcolorbox}
To enforce class-discriminative structure, we attach a classifier 
$\mathcal{C}_\phi : \mathbb{R}^k \to \mathbb{R}^{N_c}$ on top of the learned 
embedding and train it jointly with $f_\Phi$ using the total objective
\begin{equation}\label{neuralspot_total_loss}
\min_{\Phi, \phi} \underbrace{\textcolor{black}{\left\| n \pi_{\varepsilon}^*\, F_\Phi (\mathbf{X_t}) - F_\Phi (\mathbf{X_s}) \right\|_F^{2}
+ \left\| n \pi_{\varepsilon}^{*\top} F_\Phi (\mathbf{X_s}) -  F_\Phi (\mathbf{X_t})\right\|_F^{2}}}_{\text{spectral loss}}
+ \eta \underbrace{\frac{1}{n_s} \sum_{i} \mathcal{L}_{CE}\bigl(\phi(f_\Phi (\mathbf{x_s^i})), y_{s i}\bigr)}_{\text{classification loss}}
\end{equation}
where $\Phi$ and $\phi$ are optimized jointly via gradient descent, under the same setup as DoDyn.
\section{Additional Experiments}\label{app:additional_exp}
\begin{figure}[h]
\centering
\captionsetup[subfigure]{font=scriptsize,skip=1pt}
\begin{minipage}{0.8\textwidth}
\centering
\begin{subfigure}{0.19\textwidth}\includegraphics[width=\linewidth]{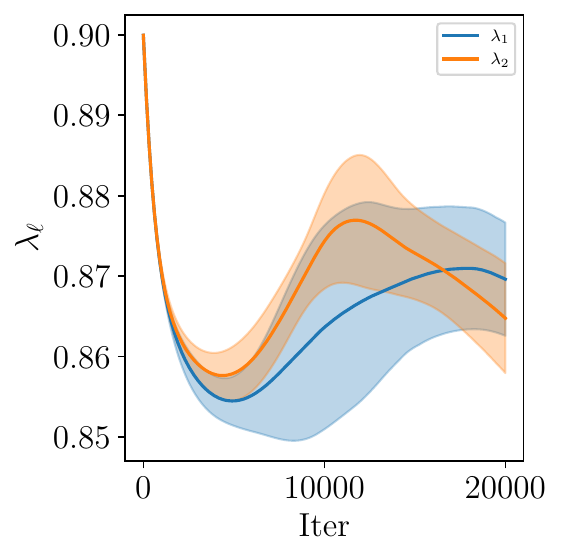}\caption{N\,$\to$\,B}\end{subfigure}\hfill
\begin{subfigure}{0.19\textwidth}\includegraphics[width=\linewidth]{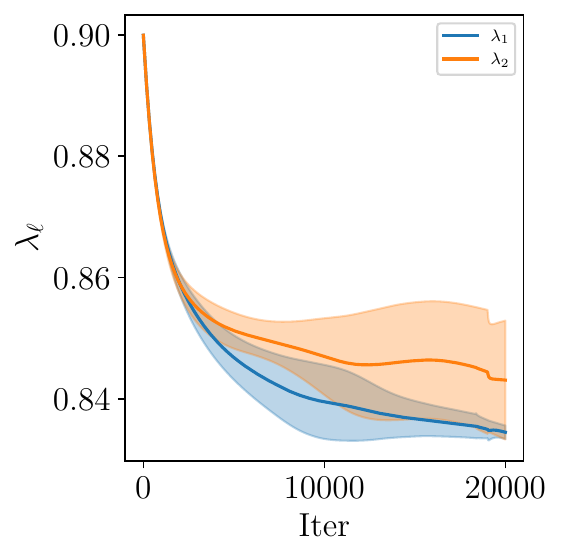}\caption{N\,$\to$\,D}\end{subfigure}\hfill
\begin{subfigure}{0.19\textwidth}\includegraphics[width=\linewidth]{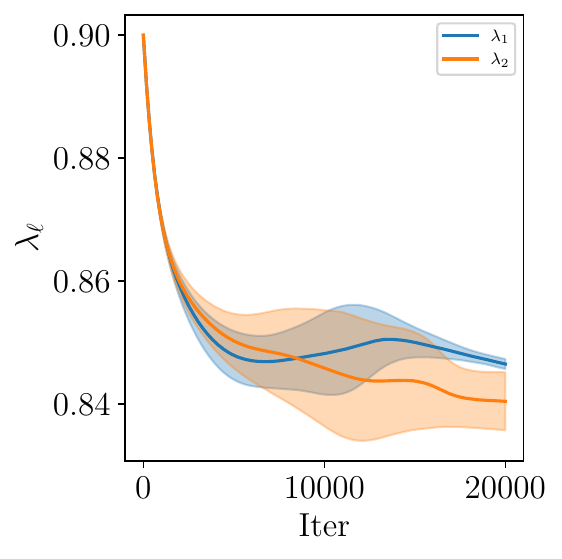}\caption{N\,$\to$\,F16}\end{subfigure}\hfill
\begin{subfigure}{0.19\textwidth}\includegraphics[width=\linewidth]{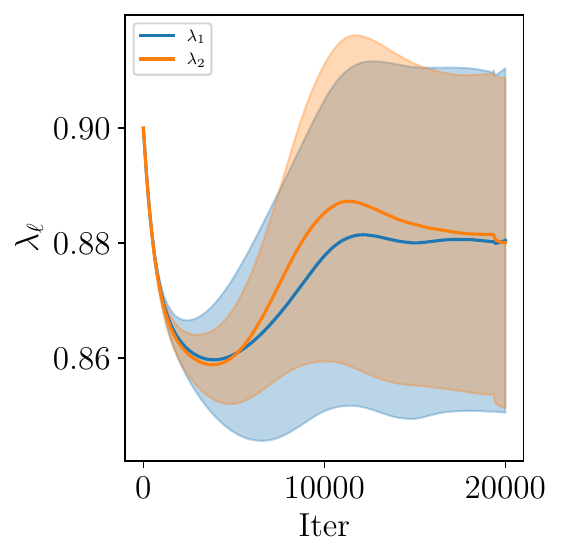}\caption{N\,$\to$\,F}\end{subfigure}\hfill
\begin{subfigure}{0.19\textwidth}\includegraphics[width=\linewidth]{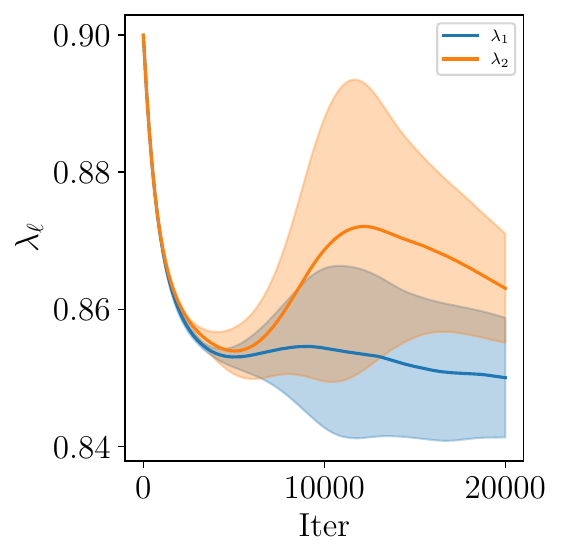}\caption{B\,$\to$\,N}\end{subfigure}
\vspace{2pt}
\begin{subfigure}{0.19\textwidth}\includegraphics[width=\linewidth]{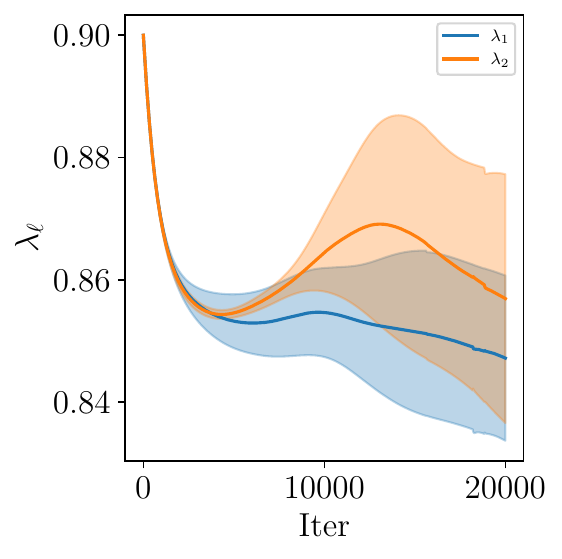}\caption{B\,$\to$\,D}\end{subfigure}\hfill
\begin{subfigure}{0.19\textwidth}\includegraphics[width=\linewidth]{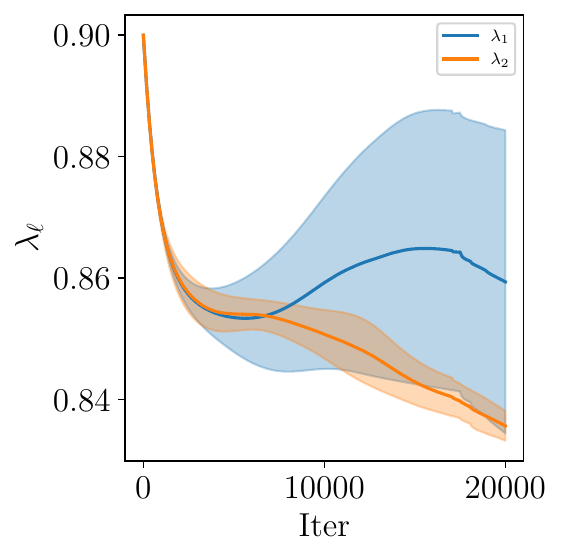}\caption{B\,$\to$\,F16}\end{subfigure}\hfill
\begin{subfigure}{0.19\textwidth}\includegraphics[width=\linewidth]{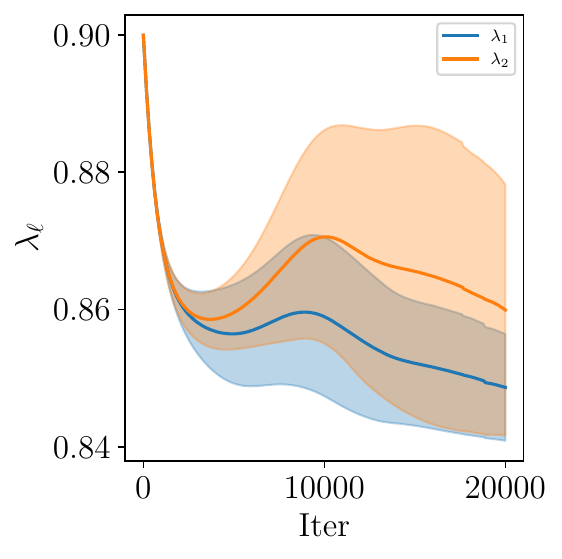}\caption{B\,$\to$\,F}\end{subfigure}\hfill
\begin{subfigure}{0.19\textwidth}\includegraphics[width=\linewidth]{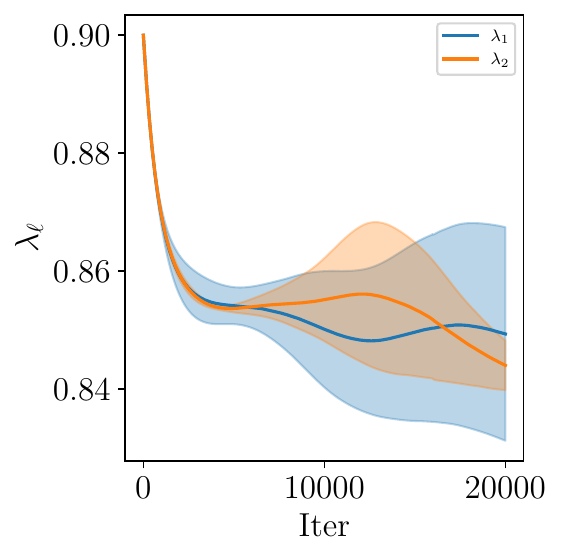}\caption{D\,$\to$\,N}\end{subfigure}\hfill
\begin{subfigure}{0.19\textwidth}\includegraphics[width=\linewidth]{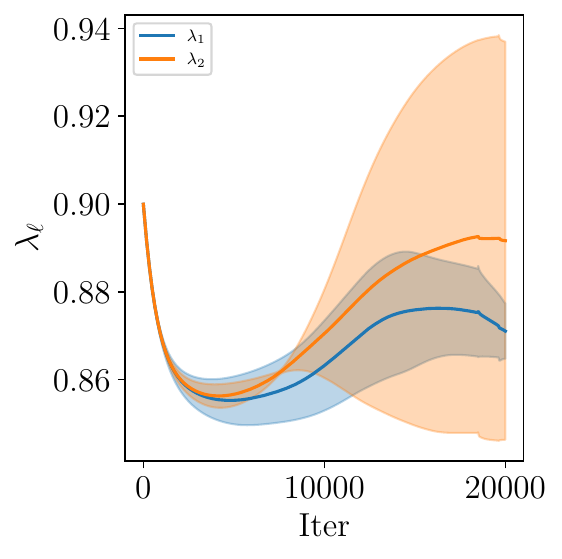}\caption{D\,$\to$\,B}\end{subfigure}
\vspace{2pt}
\begin{subfigure}{0.19\textwidth}\includegraphics[width=\linewidth]{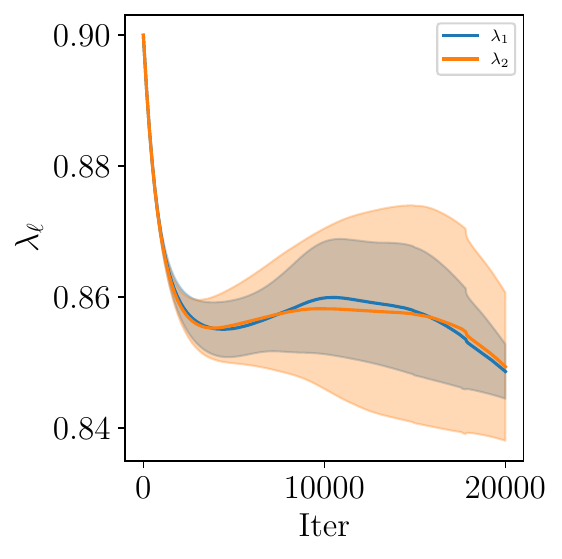}\caption{D\,$\to$\,F16}\end{subfigure}\hfill
\begin{subfigure}{0.19\textwidth}\includegraphics[width=\linewidth]{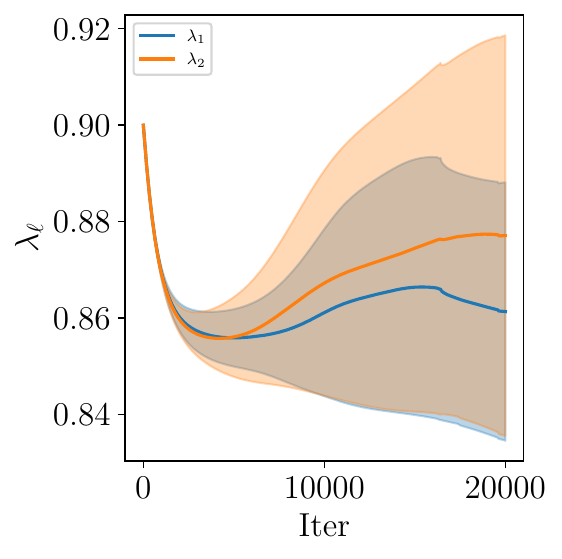}\caption{D\,$\to$\,F}\end{subfigure}\hfill
\begin{subfigure}{0.19\textwidth}\includegraphics[width=\linewidth]{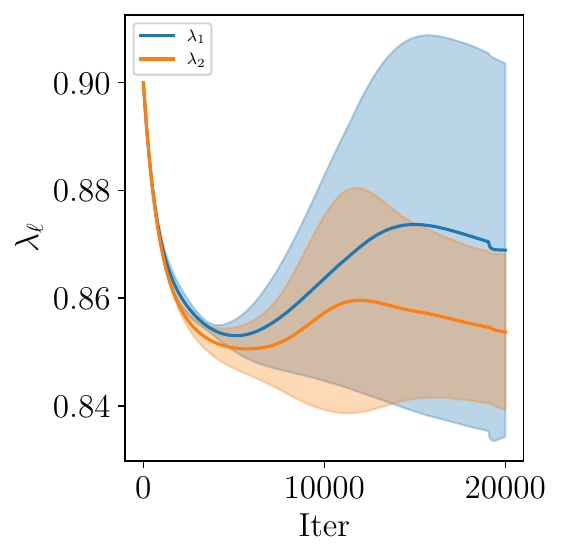}\caption{F16\,$\to$\,N}\end{subfigure}\hfill
\begin{subfigure}{0.19\textwidth}\includegraphics[width=\linewidth]{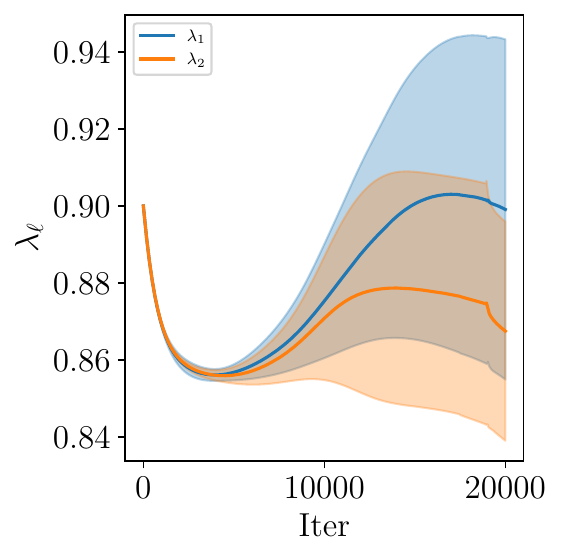}\caption{F16\,$\to$\,B}\end{subfigure}\hfill
\begin{subfigure}{0.19\textwidth}\includegraphics[width=\linewidth]{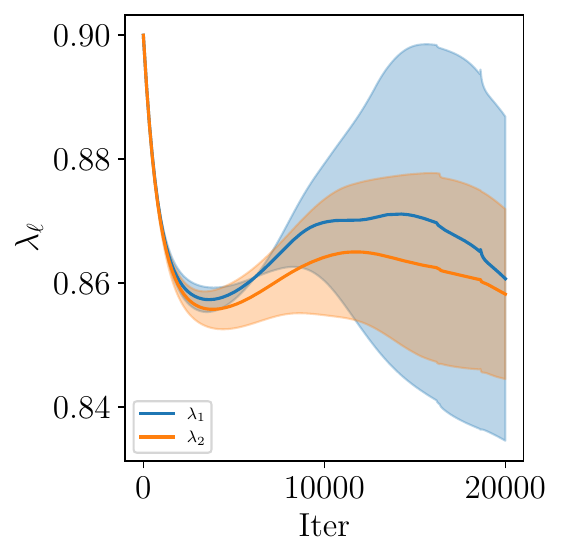}\caption{F16\,$\to$\,D}\end{subfigure}
\vspace{2pt}
\begin{subfigure}{0.19\textwidth}\includegraphics[width=\linewidth]{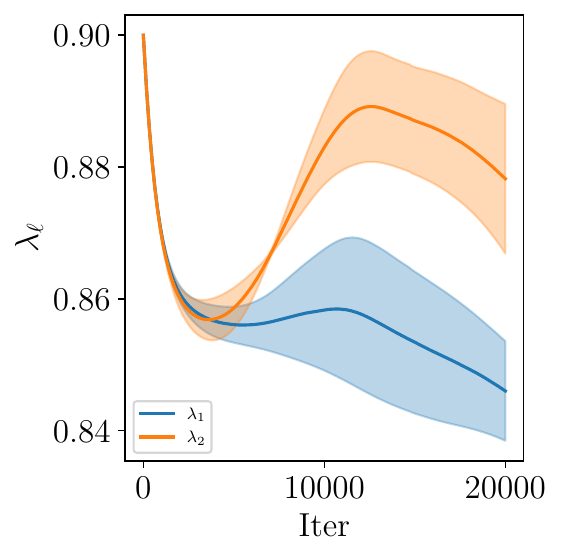}\caption{F16\,$\to$\,F}\end{subfigure}\hfill
\begin{subfigure}{0.19\textwidth}\includegraphics[width=\linewidth]{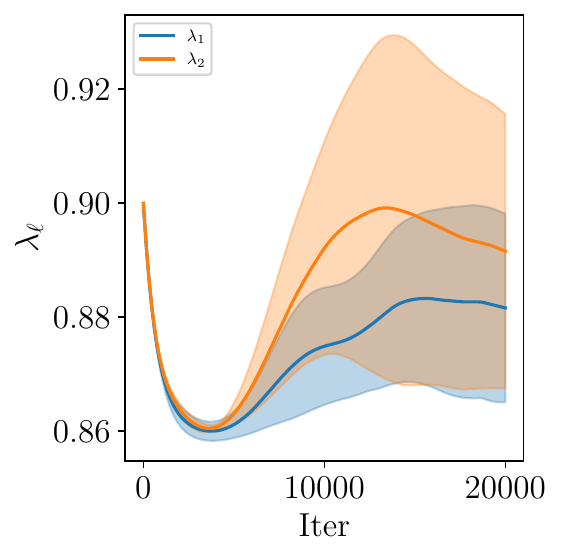}\caption{F\,$\to$\,N}\end{subfigure}\hfill
\begin{subfigure}{0.19\textwidth}\includegraphics[width=\linewidth]{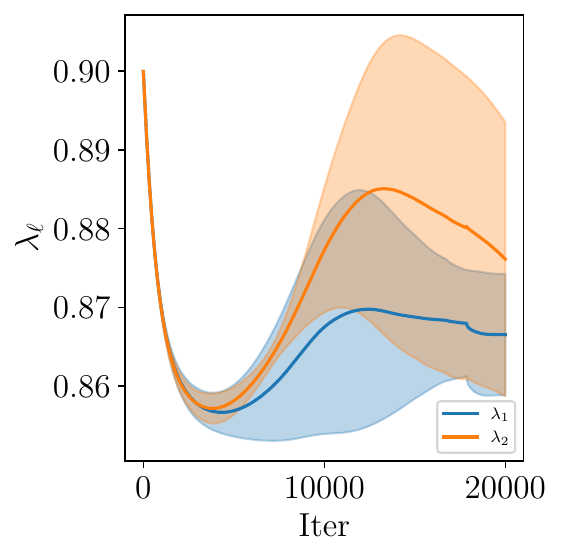}\caption{F\,$\to$\,B}\end{subfigure}\hfill
\begin{subfigure}{0.19\textwidth}\includegraphics[width=\linewidth]{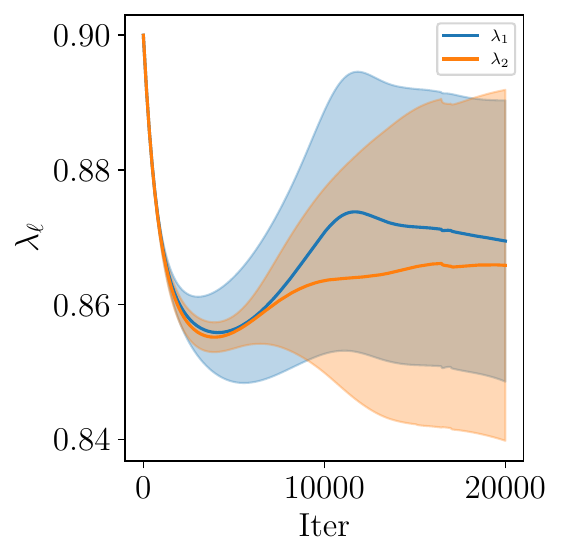}\caption{F\,$\to$\,D}\end{subfigure}\hfill
\begin{subfigure}{0.19\textwidth}\includegraphics[width=\linewidth]{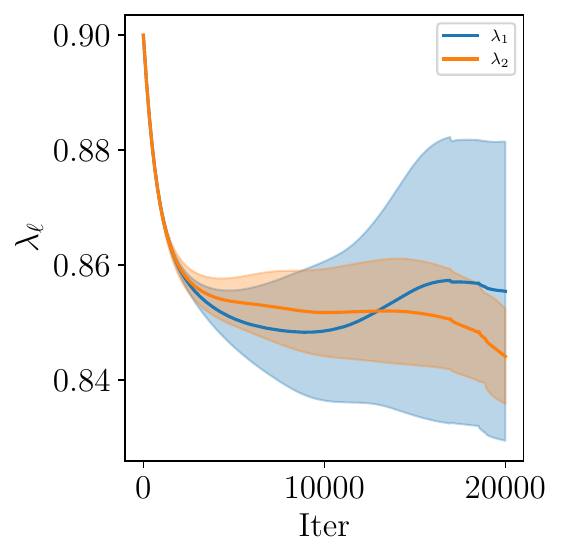}\caption{F\,$\to$\,F16}\end{subfigure}
\end{minipage}
\caption{Learned $\lambda_\ell$ updates on MSD across all source$\to$target domain pairs.}
\label{fig:lambda_msd_grid}
\end{figure}

\begin{figure}[h]
\centering
\includegraphics[width=0.365\textwidth]{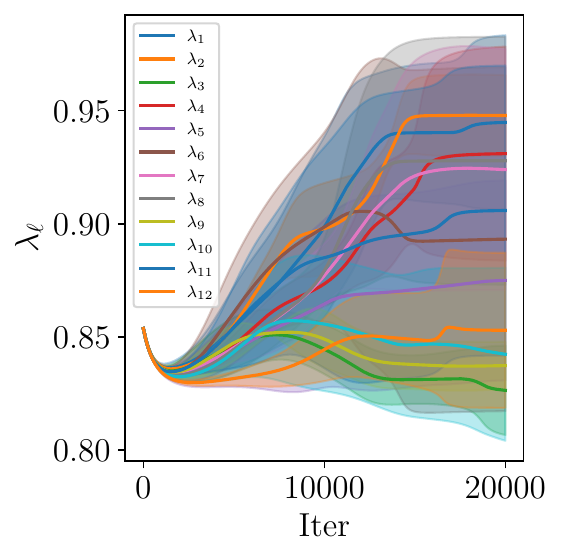}
\caption{Learned $\lambda_\ell$ updates on VisDA (synthetic\,$\to$\,real).}
\label{fig:lambda_visda}
\end{figure}

\begin{figure}[h]
\centering
\captionsetup[subfigure]{font=scriptsize,skip=1pt}
\begin{minipage}{\textwidth}
\centering
\begin{subfigure}{0.24\textwidth}\includegraphics[width=\linewidth]{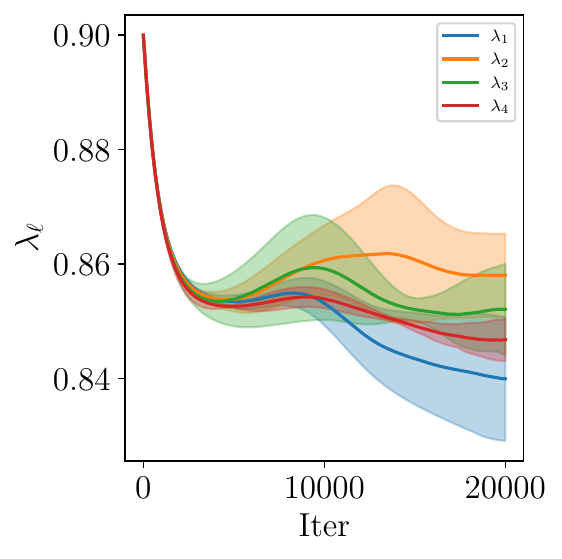}\caption{TH\,$\to$\,CF2}\end{subfigure}\hfill
\begin{subfigure}{0.24\textwidth}\includegraphics[width=\linewidth]{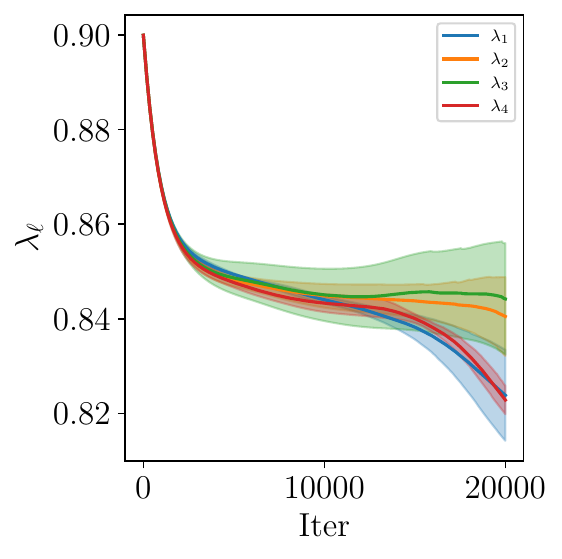}\caption{TH\,$\to$\,CF4}\end{subfigure}\hfill
\begin{subfigure}{0.24\textwidth}\includegraphics[width=\linewidth]{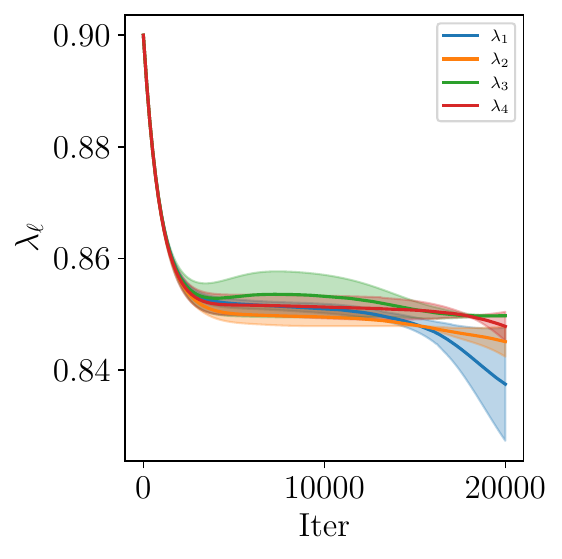}\caption{TH\,$\to$\,CF16}\end{subfigure}\hfill
\begin{subfigure}{0.24\textwidth}\includegraphics[width=\linewidth]{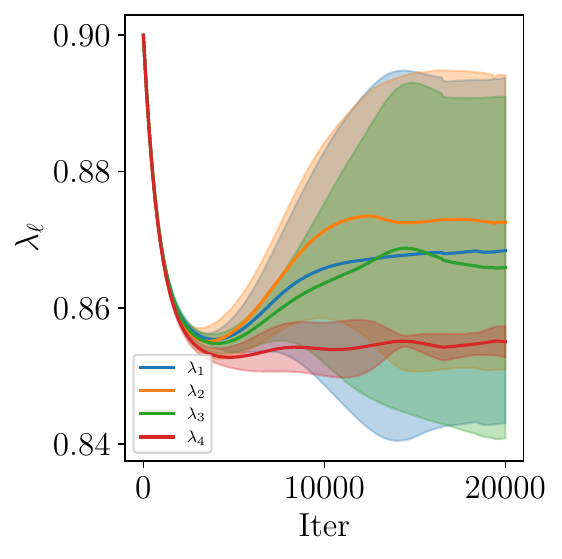}\caption{CF2\,$\to$\,TH}\end{subfigure}
\vspace{2pt}
\begin{subfigure}{0.24\textwidth}\includegraphics[width=\linewidth]{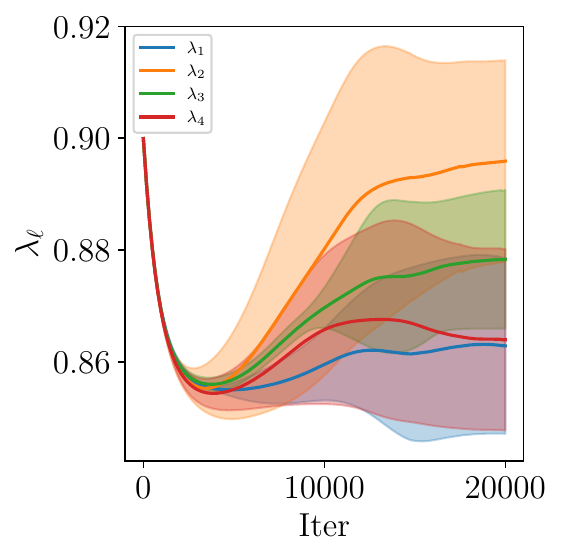}\caption{CF2\,$\to$\,CF4}\end{subfigure}\hfill
\begin{subfigure}{0.24\textwidth}\includegraphics[width=\linewidth]{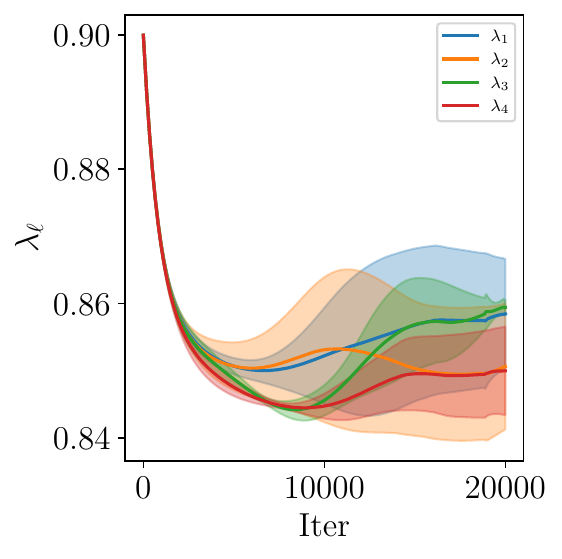}\caption{CF2\,$\to$\,CF16}\end{subfigure}\hfill
\begin{subfigure}{0.24\textwidth}\includegraphics[width=\linewidth]{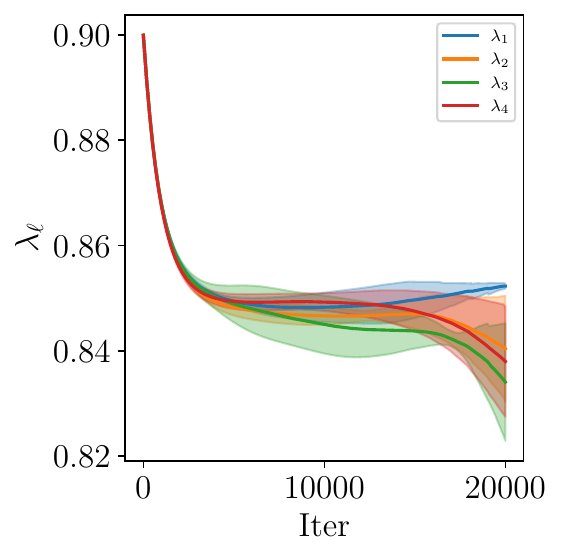}\caption{CF4\,$\to$\,TH}\end{subfigure}\hfill
\begin{subfigure}{0.24\textwidth}\includegraphics[width=\linewidth]{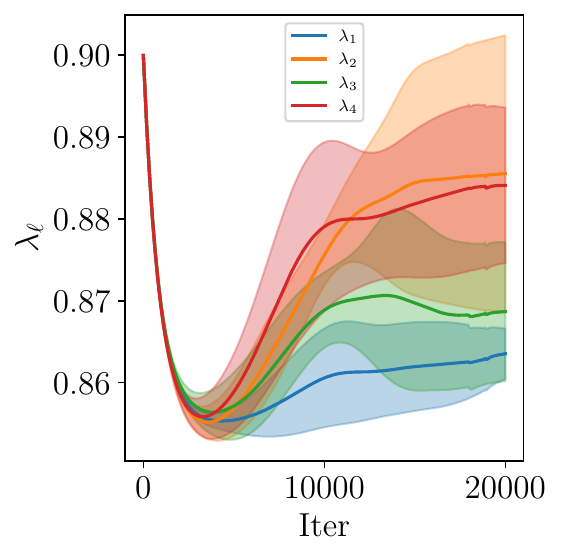}\caption{CF4\,$\to$\,CF2}\end{subfigure}
\vspace{2pt}
\begin{subfigure}{0.24\textwidth}\includegraphics[width=\linewidth]{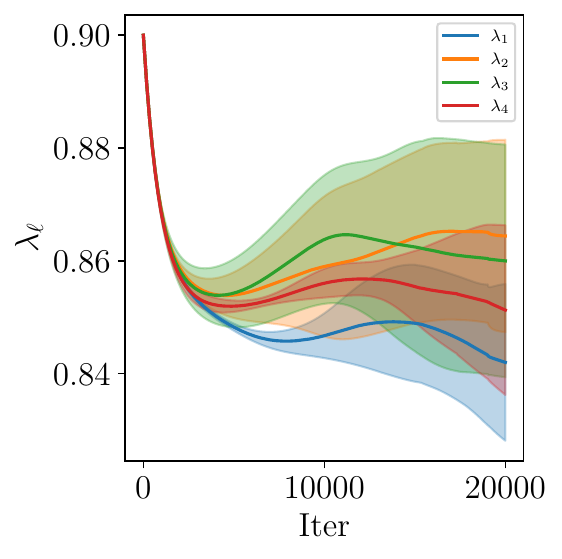}\caption{CF4\,$\to$\,CF16}\end{subfigure}\hfill
\begin{subfigure}{0.24\textwidth}\includegraphics[width=\linewidth]{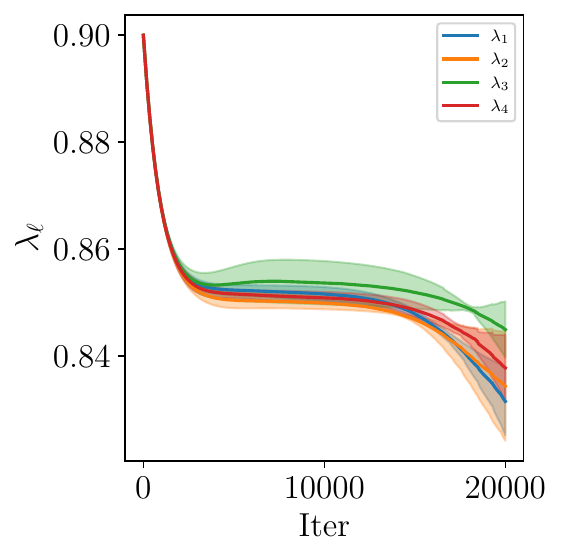}\caption{CF16\,$\to$\,TH}\end{subfigure}\hfill
\begin{subfigure}{0.24\textwidth}\includegraphics[width=\linewidth]{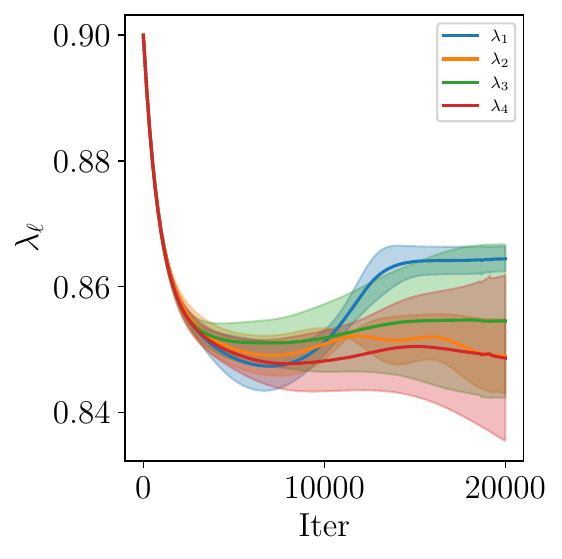}\caption{CF16\,$\to$\,CF2}\end{subfigure}\hfill
\begin{subfigure}{0.24\textwidth}\includegraphics[width=\linewidth]{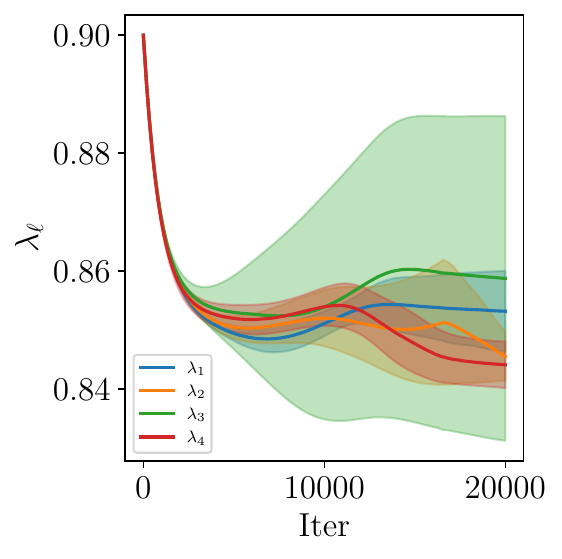}\caption{CF16\,$\to$\,CF4}\end{subfigure}
\end{minipage}
\caption{Learned $\lambda_\ell$ updates on CSRT across all source$\to$target domain pairs.}
\label{fig:lambda_csrt_grid}
\end{figure}

In this section, we present an additional study on the choice of different components of our method.
\textcolor{black}{
\subsection{On Learning the Eigenvalues $\lambda$}
\label{sec:parametrized-lambda}
A natural question is what happens when $\lambda_\ell$ is learned jointly with the other parameters, rather than fixed. To answer this, we parametrize the eigenvalue as $\lambda(\mathbf{\varphi}) = 1 - \tau\,\times\textbf{sigmoid}(\mathbf{\varphi})$, where $\mathbf{\varphi}$ is a trainable parameter added alongside $\boldsymbol{\Theta}$ and $\phi$, and $\tau$ controls how far below 1 the eigenvalue can move, which we set to $\tau = 0.2$ \textit{w.r.t} $\lambda \lesssim 1$ throughout training. Under this setting, the objective \eqref{dodyn_objective} then becomes:
\begin{align}
\underset{\boldsymbol{\Theta},\,\phi,\,\mathbf{\varphi}}{\operatorname{argmin}}
\quad
\left\{
\begin{aligned}
\bigl\|\boldsymbol{\Theta}\operatorname{diag}\!\big(\lambda(\mathbf{\varphi})\big) - \mathbf{M_\pi}\boldsymbol{\Theta}\bigr\|_F^2
+\;&\eta\,\mathcal{L}_{\mathrm{CE}}\bigl(\mathcal{C}_\phi(\hat f_{\boldsymbol{\Theta}}(\mathbf{X_s})), \mathbf{Y_s}\bigr)
\end{aligned}
\right\}
\label{full_objective}
\end{align}}
\begin{wraptable}{r}{0.4\textwidth}
\centering
\caption{Comparison with learned eigenvalue $\lambda_\ell$ extension of DoDyn across benchmarks.}
\label{tab:results}
\resizebox{0.4\textwidth}{!}{%
\begin{tabular}{l cc cc cc}
\toprule
& \multicolumn{2}{c}{MSD} & \multicolumn{2}{c}{CSRT} & \multicolumn{2}{c}{VisDA} \\
\cmidrule(lr){2-3} \cmidrule(lr){4-5} \cmidrule(lr){6-7}
Algorithm & $\mathcal{A}_T$ & $\mathcal{A}_{\mathrm{OOS}}$ & $\mathcal{A}_T$ & $\mathcal{A}_{\mathrm{OOS}}$ & $\mathcal{A}_T$ & $\mathcal{A}_{\mathrm{OOS}}$ \\
\midrule
Source-only & 85.25 & 85.25 &37.35 &37.66 & 54.69& 55.14 \\
DoDyn$_{\lambda}$    & \underline{92.68} & \underline{91.45}& \underline{53.04}&\underline{61.61} &\underline{69.16} & \underline{69.37}\\
DoDyn              & \textbf{94.73} & \textbf{94.63}& \textbf{58.49} & \textbf{64.74}&\textbf{69.71} & \textbf{69.81} \\
\bottomrule
\end{tabular}}
\end{wraptable}
\textcolor{black}{We refer to this extension as DoDyn$_\lambda$. The learned eigenvalue is a vector of dimension $k$, so that each eigenfunction $f_\ell$ is paired with its own eigenvalue $\lambda_\ell$. We compare DoDyn against DoDyn$_\lambda$ on MSD ($k=2$), CSRT ($k=4$), and VisDA ($k=12$), reporting both the target accuracy $\mathcal{A}_T$ and the out-of-sample accuracy $\mathcal{A}_{\mathrm{OOS}}$ in Table~\ref{tab:results}, showing that both variants improve clearly over the source-only baseline across all three benchmarks, while DoDyn (with fixed $\lambda$) stays ahead of DoDyn$_{\lambda}$, by a slight margin on some benchmarks. We also track how the learned eigenvalues $\lambda$ evolve during training, for every domain pair and across multiple seeds. These trajectories are shown in Figure~\ref{fig:lambda_msd_grid} for MSD, Figure~\ref{fig:lambda_csrt_grid} for CSRT, and Figure~\ref{fig:lambda_visda} for VisDA. }


\subsection{Monte Carlo sample budget and computational cost}

\begin{figure}[h]
\centering
\begin{subfigure}[h]{0.3\textwidth}
\centering
\includegraphics[width=\linewidth]{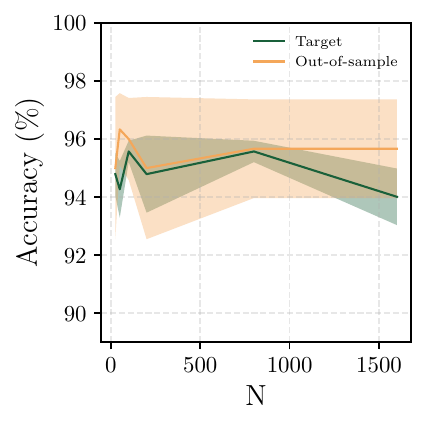}
\caption{}
\label{fig:ablation_mcs}
\end{subfigure}
\begin{subfigure}[h]{0.3\textwidth}
\centering
\includegraphics[width=\linewidth]{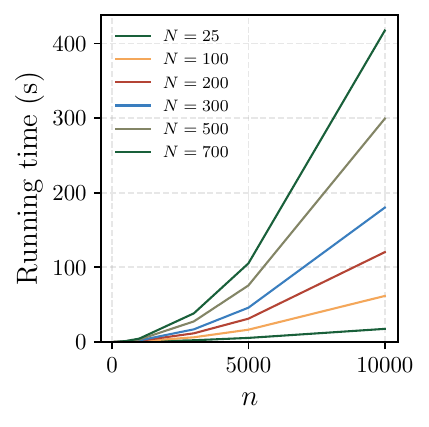}
\caption{}
\label{fig:ablation_timing}
\end{subfigure}
\caption{(a) Sensitivity of DoDyn to the Monte Carlo sample budget $N$ used to estimate the overlap matrices, evaluated on the MSD benchmark (Noiseless $\rightarrow$ Buccaneer). (b) Running time for computing the four overlap matrices as a function of the number of samples $n$, for varying $N$, on synthetic data.}
\label{fig:ablation_mc}
\end{figure}

We examine the effect of the number of Monte Carlo samples $N$ on the adaptation performance and the associated computational cost of the overlap matrices. In Fig.~\ref{fig:ablation_mcs}, we vary $N$ over $\mathbf{\{25, 50, 100, 200, 800, 1600\}}$ on the MSD benchmark (N $\rightarrow$ B) and report target and out-of-sample accuracies, 
with all other hyperparameters fixed to their default values. In Fig.~\ref{fig:ablation_timing}, we report the wall-clock time required to compute 
the four overlap matrices on synthetic data of increasing size $n \in \mathbf{\{100, 200, 500, 1000, 3000, 5000, 10000\}}$, where 
source and target samples are drawn from Gaussian distributions, for five values of 
$N \in \mathbf{\{25, 100, 200, 300, 500, 700\}}$. As shown in Fig.~\ref{fig:ablation_mcs}, both target and out-of-sample accuracy remain almost stable across the full range of $N$, which indicates that the Monte Carlo estimators of the overlap matrices converge rapidly and that a modest $N$ suffices in practice. Concerning the runtime in Fig.~\ref{fig:ablation_timing}, it 
grows with $n$ for all values of $N$, consistent with the $\mathcal{O}(n^2 N d)$ complexity of the estimation step.

\newpage
\subsection{DoDyn's Components Contribution}

\begin{figure}[t]
\centering
\begin{subfigure}[b]{0.3\textwidth}
\centering
\includegraphics[width=\linewidth]{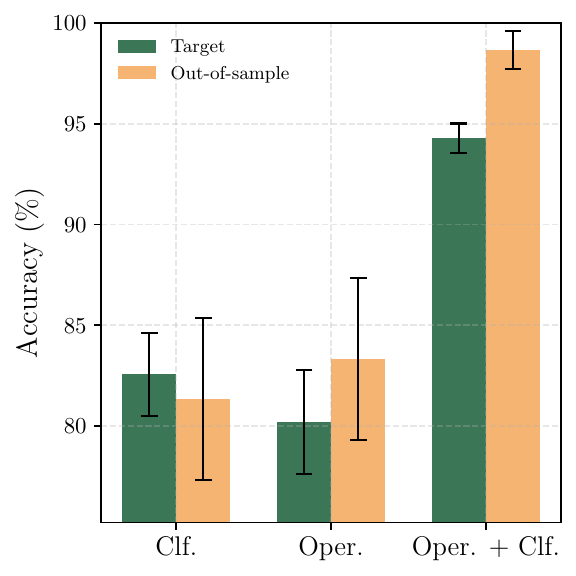}
\label{fig:ablation_structural_sigma}
\end{subfigure}
\hspace{0.03\textwidth}
\begin{subfigure}[b]{0.3\textwidth}
\centering
\includegraphics[width=\linewidth]{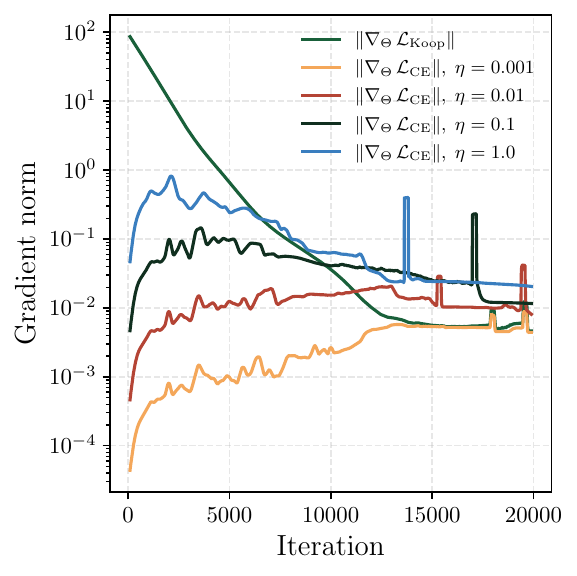}
\label{fig:ablation_grad_eta}
\end{subfigure}
\caption{Ablation on DoDyn's structural components. \textit{(Left)} Accuracy of five configurations. \textit{(Right)} Gradient norms throughout training for varying $\eta$.}
\label{fig:ablation_structural}
\end{figure}

To study the contribution of each component of \textsc{DoDyn}, we compare three configurations on the \textsc{MSD} benchmark using the adaptation task Noiseless~$\rightarrow$~buccaneer2, with all hyperparameters fixed to their default values (isotropic lifting, $\sigma = 0.1$, $\varepsilon = 0.001$, $N = 200$, $k = 2$), and report both target and out-of-sample accuracies. The configurations are:
\textit{(i)}~\textit{Clf.}, a classifier trained on source samples and evaluated at target domain with no adaptation. \textit{(ii)}~\textit{Oper.}, a learned $\Theta$ via $\|\boldsymbol{\Theta}_{\lambda_\ell}-\mathbf{M_\pi}\boldsymbol{\Theta}_{\lambda_\ell}\|_F^2$ only without the classification term, once $\Theta$ is obtained, a post-hoc classifier is trained on the source embeddings and evaluated on target/OOS ones. \textit{(iii)}~\textit{Oper.~+~Clf.}, which implements the proposed \textsc{DoDyn} objective. Fig.~\ref{fig:ablation_structural} shows that the proposed \textsc{DoDyn} objective (\textit{Oper.~+~Clf.}) achieves the best performance, with \textit{Oper.} and \textit{Clf.} performing almost comparably, confirming the importance of the combination in~\eqref{dodyn_objective} with the classifier contribution. The gradient analysis in Fig.~\ref{fig:ablation_structural} (right) further supports the design of \eqref{dodyn_objective}, where $\|\nabla_{\Theta}\,\mathcal{L}_{\mathrm{Koop}}\|$ dominates $\|\nabla_{\Theta}\,\mathcal{L}_{\mathrm{CE}}\|$ by one to two orders of magnitude throughout training, meaning the classifier helps learning the class-discriminative spectral embedding rather than distorting it. This also provides a practical guideline to select $\eta$ in a way that offers a good balance, as larger values risk pushing the \textsc{CE} gradient too close to the Koopman gradient and perturbing the solution.

\subsection{Experimental setup and hyperparameters}
\label{app:experimental_setup}
For each dataset we keep 80\% of the target domain to compute the transport plan and reserve the remaining 20\% for out-of-sample evaluation, except for CSRT where we instead generate 200 samples independent from the target set for out-of-sample evaluation.

We model each domain as a Gaussian mixture following the protocol of \citep{montesuma2025optimal}, with all parameters estimated by Expectation-Maximization (EM) \citep{dempster1977maximum}. The labeled source is fit class-conditionally, while the unlabeled target is fit by standard EM. The isotropic lifting used elsewhere is the special case in which every sample is lifted to represent its own component, i.e. means fixed at the samples with a shared fixed covariance $\sigma^2 I$, so that no EM step is required.

The algorithm is run on the full source and target domains for $n_{\mathrm{iter}}=20000$ iterations,  the optimal values of the hyperparameters $\varepsilon$, $\sigma$, $N$, and $\eta$ are discussed in the ablation study, and the number of eigenfunctions $k$ is always set equal to the number of classes $N_c$. 

For MSD, MGR, CSRT, and CMU-PIE no feature extraction procedure is applied, and the classifier is a multilayer perceptron with a hidden layer of size 1024 followed by a ReLU activation and a linear output layer, trained with Adam at a learning rate of $10^{-3}$ for 100 epochs in the other baselines using minibatches of size 128, whereas in DoDyn it is trained end to end jointly with the spectral coordinates for the same number of $n_{iter}$, with no fine-tuning afterward. For VisDA we follow the same feature extraction procedure as \cite{montesuma2025optimal}, using a ViT backbone whose details we refer the reader to therein, and the classifier is a single linear layer trained with batch size 128 (for compared baselines), SGD at a learning rate of $10^{-4}$ and a momentum = 0.9.

\end{document}